\documentclass{article}

\usepackage[preprint,nonatbib]{nips_2018}

\usepackage[utf8]{inputenc} 
\usepackage{hyperref}       
\usepackage{url}            
\usepackage{booktabs}       
\usepackage{amsfonts}       
\usepackage{nicefrac}       
\usepackage{microtype}      

\usepackage{textcomp}
\usepackage{xspace}
\usepackage{amsmath}
\usepackage[linesnumbered, inoutnumbered]{algorithm2e}
\usepackage{xcolor}
\usepackage{dsfont}
\usepackage{amssymb}
\usepackage{graphicx}
\usepackage{subfig}

\newcommand{\RR}{\mathbb{R}} 
\newcommand{\half}[0]{\frac{1}{2}}

\newcommand{\probi}[2]{\ensuremath{\mathbb{P}^{#1}({#2})}}
\newcommand{\probs}[1]{\ensuremath{\mathbb{P}^{#1}}}

\newcommand{\expectv}{\mathbb{E}}
\newcommand{\indic}{\mathds{1}}

\newcommand{\down}[0]{\emph{down}\xspace}
\newcommand{\up}[0]{\emph{up}\xspace}

\title{Deep PDF: Probabilistic Surface Optimization and Density Estimation}

\author{
	Dmitry Kopitkov\thanks{Dmitry Kopitkov is with the Technion Autonomous Systems Program (TASP), {\tt dimkak@technion.ac.il}.} 
	\And \\
	Technion - Israel Institute of Technology\\
	Haifa 32000, Israel
	\And 
	Vadim Indelman\thanks{Vadim Indelman is with the Department of Aerospace Engineering, {\tt vadim.indelman@technion.ac.il}.}
}

\begin{document}
	
	\maketitle
	
	\begin{abstract}
		
		A probability density function (pdf) encodes the entire stochastic knowledge about data distribution, where data may represent stochastic observations in robotics, transition state pairs in reinforcement learning or any other empirically acquired modality. Inferring data pdf is of prime importance, allowing to analyze various model hypotheses and perform smart decision making. However, most density estimation techniques are limited in their representation expressiveness to specific kernel type or predetermined distribution family, and have other restrictions. For example, kernel density estimation (KDE) methods require meticulous parameter search and are extremely slow at querying new points. In this paper we present a novel non-parametric density estimation approach, DeepPDF, that uses a neural network to approximate a target pdf given samples from thereof. Such a representation provides high inference accuracy for a wide range of target pdfs using a relatively simple network structure, making our method highly statistically robust. This is done via a new stochastic optimization algorithm, \emph{Probabilistic Surface Optimization} (PSO), that turns to advantage the stochastic nature of sample points in order to force network output to be identical to the output of a target pdf. Once trained, query point evaluation can be efficiently done in DeepPDF by a simple network forward pass, with linear complexity in the number of query points. Moreover, the PSO algorithm is capable of inferring the frequency of data samples and may also be used in other statistical tasks such as conditional estimation and distribution transformation. We compare the derived approach with KDE methods showing its superior performance and accuracy.

	\end{abstract}

	\section{Introduction}
	\label{sec:Intro}

	Density estimation is a fundamental statistical problem which is essential for many scientific fields and application domains such as robotics, AI, economics, probabilistic planning and inference. Today's  most statistically robust techniques that work on a wide range of probabilistic distributions are non-parametric where arguably the leading is kernel density estimation (KDE). Though a large amount of research was done on these techniques, they still are computationally expensive at both estimation and query stages. They require searching for hyper-parameters like bandwidth and kernel type, and their expressiveness in representing the estimated probability density function (pdf) is high but still limited to the selected kernel. On the other hand, neural networks (NN) are known for their high flexibility and ability to approximate any function. Moreover, recently strong  frameworks (e.g.~\cite{Tensorflow_url, Pytorch_url, Caffe_url}) were developed that allow fast and sophisticated training of NNs using GPUs. 
	
	With this motivation, 
	in this paper we present a novel method to perform non-parametric density estimation using Deep Learning (DL) where we exploit the approximation power of NN in full. Unlike any existing density estimation techniques, our approach, named DeepPDF, estimates a target pdf from sampled data using our novel stochastic optimization that forces NN's output to be identical to the output of a target pdf function, thereby approximating the latter by a NN.

	The KDE method is one of the most expressive methods that is flexible enough to approximate a broad diapason of target pdfs. It represents the estimated pdf through a sum of kernel functions, with kernel for each sample point. A naive implementation of such an approach has complexity of $O(Nm)$, where $N$ is number of sample points and $m$ is number of query points. Although faster KDE extensions exist (e.g. \cite{OBrien16csda}), they are still relatively slow. Specifically, query stage of state-of-the-art fastKDE method \cite{OBrien16csda} has quasi-linear time complexity $O(m \log m)$. Additionally, most KDE methods require to specify kernel type and bandwidth for better performance. Optimal values of these parameters are pdf-specific and parameter fine-tuning needs to be done. 
	Moreover, the target pdf may have a complex structure with different parts requiring different optimal KDE parameters. In such case, standard KDE methods will return a statistically sub-optimal solution.

\begin{figure}[t]
		\centering
		
		\begin{tabular}{ccc}
			
			\subfloat[\label{fig:Overview-a}]{\includegraphics[width=0.45\textwidth]{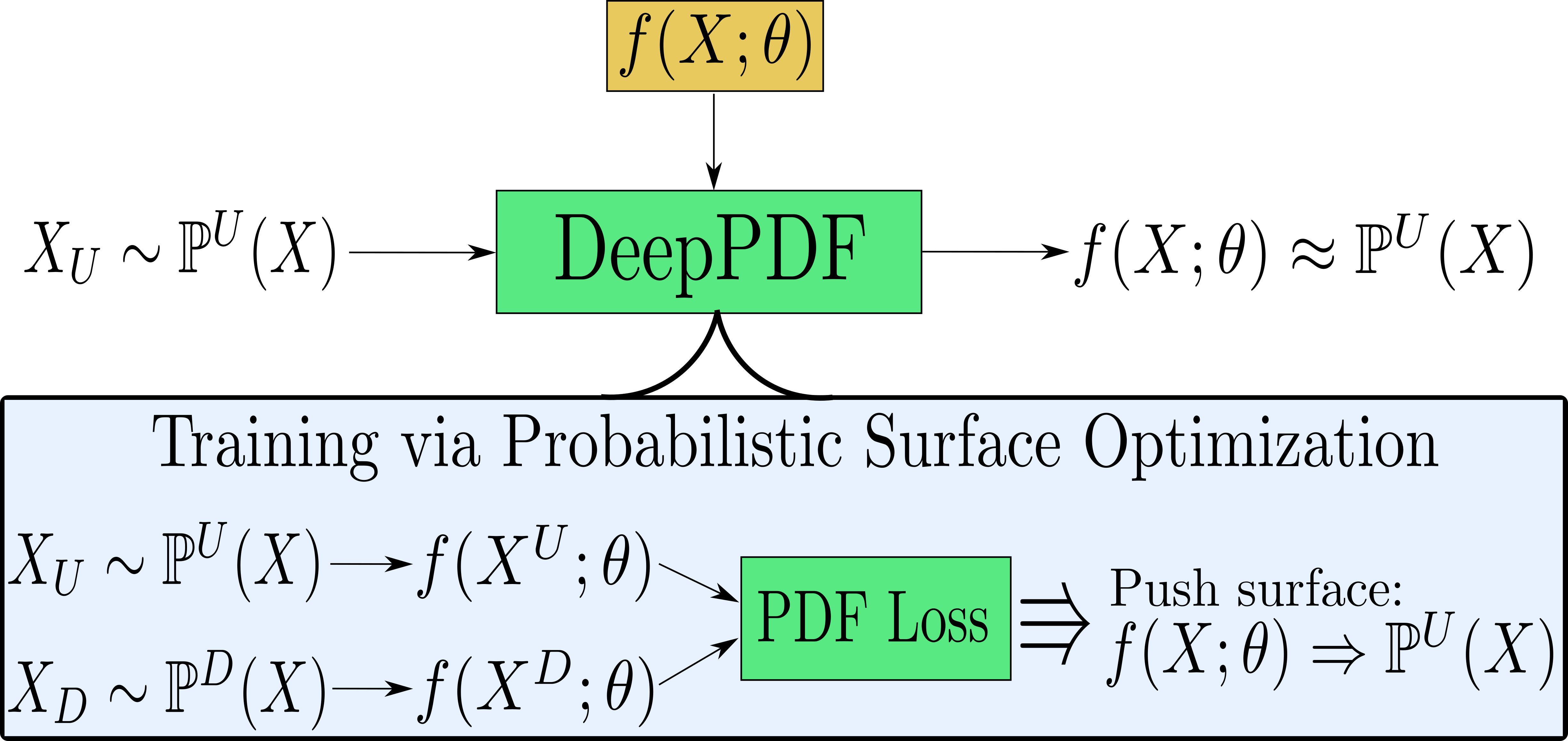}}
			&
			
			\subfloat[\label{fig:Overview-b}]{\includegraphics[width=0.23\textwidth]{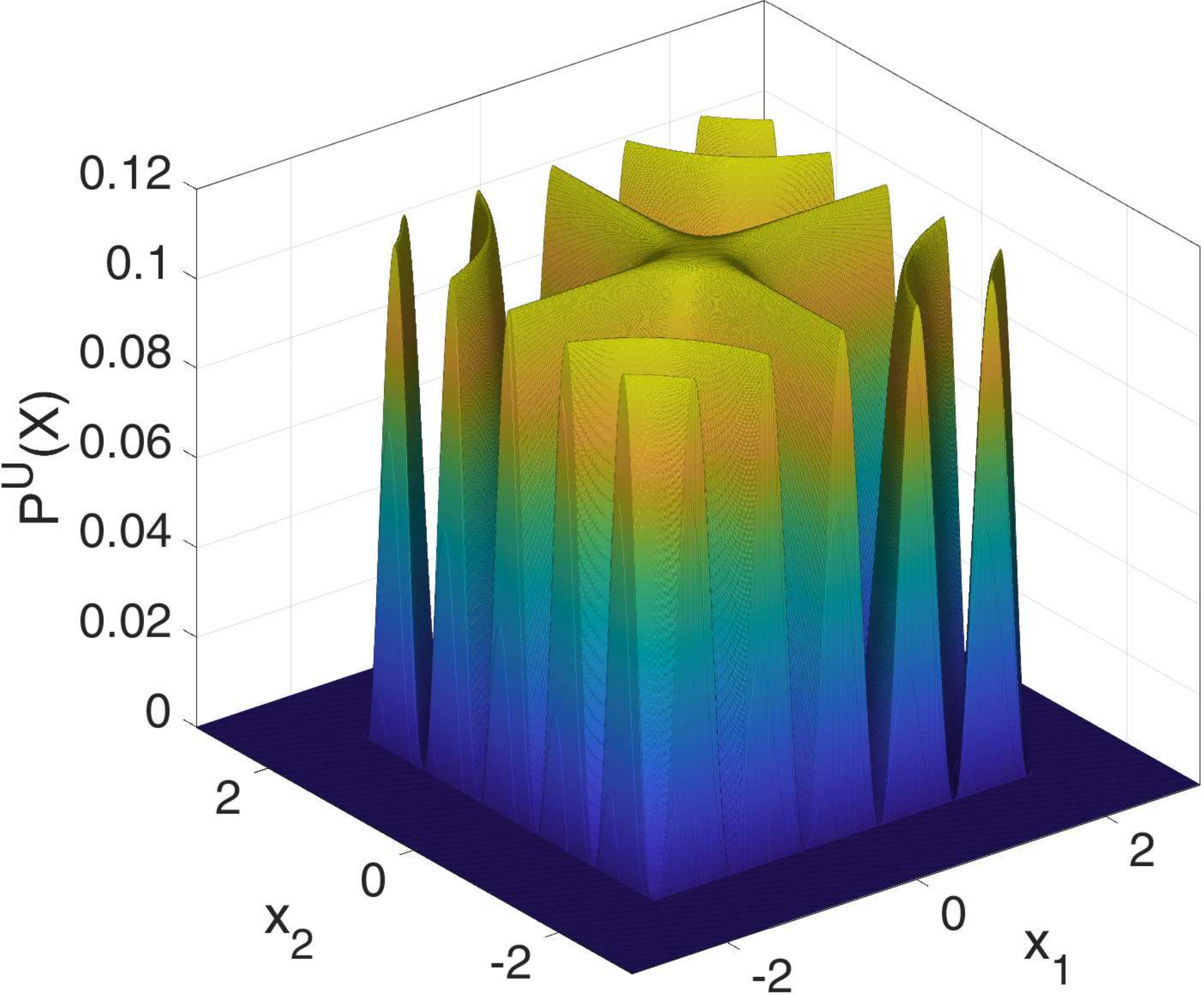}}
			&		
			
			\subfloat[\label{fig:Overview-c}]{\includegraphics[width=0.23\textwidth]{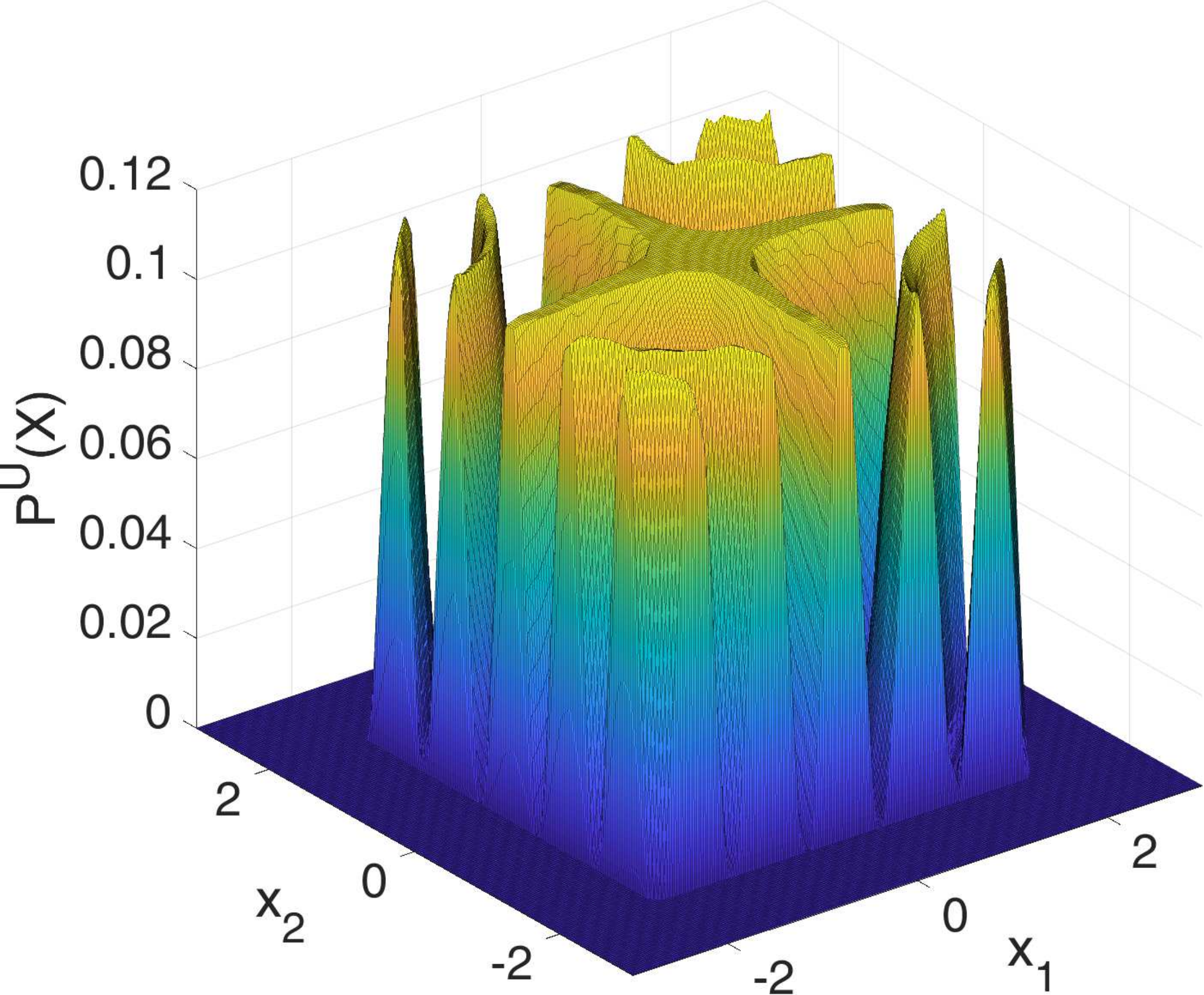}}
			
		\end{tabular}
		
		\protect
		\caption{(a) Approach overview. Our method, DeepPDF, receives data samples $X_U \sim \probi{U}{X}$ from an arbitrary unknown pdf $\probi{U}{X}$ and optimizes a neural network $f(X; \theta)$ with scalar output to approximate the pdf $\probi{U}{X}$. 
			This is done using a novel \emph{Probabilistic Surface Optimization} (PSO) and \emph{pdf loss}. The PSO uses target samples $X_U \sim \probi{U}{X}$ and samples from some known distribution $X_D \sim \probi{D}{X}$ to push surface $f(X; \theta)$ at these sample points, making it eventually identical to target $\probi{U}{X}$. (b) Example \emph{Cosine} target pdf $\probi{U}{X}$ and (c) its DeepPDF estimation.
		}
		\label{fig:Overview}
	\end{figure}

	In contrast, the presented herein DeepPDF approach (see also Figure \ref{fig:Overview-a}) allows to approximate the target pdf via a NN of arbitrary architecture and takes advantage of approximation robustness embedded within DL methods. Our approach trains a NN using a \emph{Probabilistic Surface Optimization} (PSO). The PSO, being one of our main contributions, turns to advantage the stochastic nature of each data sample point in order to push surface that is represented by the NN. Likewise, this surface is pushed at points sampled from another known distribution. We use specially discovered stochastic \emph{pdf loss} to push the neural surface in the opposite directions at data samples and at known distribution samples, where balance (minimum of \emph{pdf loss}) is achieved only when the NN surface is identical to the target density. Such a balance is a substantial attribute of \emph{pdf loss} which implicitly forces the estimator function (the surface) to have total integral of 1 and to be non-negative, releasing us from intricate dealing with these constraints explicitly. Moreover, training DeepPDF can be done in batch mode where at each iteration only a small number of sample points is used, lowering the requirement on memory size.
	DeepPDF is highly expressive and able to use facilities and theory developed lately for efficient and accurate DL training, including fast-converging optimizers, regularization techniques and execution on GPU cards. The pdf value at query point can be calculated by just computing output of trained NN with query point as input, which can be done very fast with complexity only depending on the NN's size, in contrast to quasi-linear complexity of fastKDE query stage.

	To summarize, our main contributions in this paper are as follows: 
	(a) we develop a \emph{Probabilistic Surface Optimization} (PSO) and \emph{pdf loss} that force any approximator function to converge to the target density;
	(b) we use PSO to approximate a target density using a NN;
	(c) we train a NN via PSO in batch mode using exponential learning rate decay for better convergence, naming the entire algorithm DeepPDF;
	and (d) we analyze different DL aspects of DeepPDF.
	Additionally, for simplicity the derivations are gathered together under Appendix, at the end of the paper.

	\section{Related work}
	\label{sec:RelW}

	Statistical density estimation may be divided into two different branches - parametric and non-parametric. Parametric methods assume data to come from a probability distribution of a specific family, and infer parameters of that family, for example via minimizing the negative log-probability of data samples. Non-parametric approaches are distribution-free in the sense that they do not take any assumption over data population a priori. Instead they infer the distribution density totally from  data.

	The main advantage of parametric approaches is their statistical efficiency. Given the assumption of a specific distribution family is correct, parametric methods will produce more accurate density estimation for the same number of samples compared to non-parametric techniques. 
	However, as was also observed in our previous work \cite{Kopitkov18iros}, in case the assumption is not entirely valid for a given population, the estimation accuracy will be poor, making parametric methods not statistically robust. For example, one of the most flexible distribution families is a Gaussian Mixture Model (GMM) \cite{McLachlan88}. One of its structure parameters is the number of mixtures. Using a high number of mixtures, it can represent multi-modal populations with high accuracy. Yet, in case the real unknown distribution has even higher number of modes, or sometimes even an infinite number (see Section \ref{sec:Exper3D} and Figure \ref{fig:Res2-c}), the performance of a GMM will be low.

	To handle the problem of unknown number of mixture components in parametric techniques, Bayesian statistics can be applied to model a prior over parameters of the chosen family. Models such as Dirichlet process mixture (DPM) and specifically Dirichlet process Gaussian mixture model (DPGMM) \cite{Antoniak74aos, Sethuraman82sdtrt, Gorur10jcst} can represent uncertainty about the learned distribution parameters and as such can be viewed as infinite mixture models. Although such hierarchical models are more statistically robust (flexible), they still require to manually select a base distribution for DPM, limiting its robustness. Also, Bayesian inference applied in these techniques is more theoretically intricate and computationally expensive \cite{MacEachern98}.

	On the other hand, non-parametric approaches can infer distributions of arbitrary form. Methods such as data histogram and kernel density estimation (KDE) \cite{Scott15book, Silverman18} use frequencies of different points within data samples in order to conclude how a population pdf looks like. In general, these methods require more samples but also provide more robust estimation by not taking any prior assumptions. Both histogram and KDE require selection of parameters - bin width for histogram and kernel type/bandwidth for KDE which in many cases require manual parameter search \cite{Silverman18}. Although an automatic parameter deduction was proposed in several studies \cite{Duong05sjs, Heidenreich13asta, OBrien16csda}, it is typically computationally expensive and its performance is not always optimal. Moreover, one of the major weaknesses of the KDE technique is its time complexity during query stage. Even the most efficient KDE methods (e.g. fastKDE \cite{OBrien16csda}) require above linear complexity ($O(m \log m)$) in the number of query points $m$. In contrast, the method presented herein, DeepPDF, is a robust non-parametric approach that calculates the pdf value of a query point by a single forward pass of a NN, making query complexity to be linear in $m$. Such query complexity supremacy makes DeepPDF a much better alternative when many query points require evaluation. Moreover, as we will see in Section \ref{sec:Exper}, DeepPDF provides comparable and sometimes even a much higher approximation accuracy.

	Using NNs for density estimation was studied for several decades \cite{Smolensky86rep, Bishop94tr, Bengio00nips, Hinton06, Uria2013nips}. The Mixture Density Network (MDN) presented by \cite{Bishop94tr} represents parameters of a GMM through a NN, taking the advantage of its flexibility and expressiveness. Additionally, Williams at el. \cite{Williams96nc} introduced an efficient representation to parametrize a positive-definite covariance matrix of a Gaussian distribution through a NN. In image classification CNN networks produce discrete class probabilities \cite{Krizhevsky12nips} for each image. Yet, most of the work was done in the context of parametric approaches, arguably, due to their relative analytical simplicity.

	Another body of research in DL-based density estimation was explored
	in \cite{Larochelle11aistats, Uria2013nips, Germain15icml}. The described autoregressive methods NADE, RNADE and MADE decompose the joint distribution of a  multivariate data into a product of simple conditional densities where a specific variable ordering needs to be selected for better performance. Although these approaches provide high statistical robustness, NADE and MADE are purposed to infer density of discrete binary multi-dimensional data which limits their applicability to real world data. Further, RNADE was developed to overcome this limitation, inferring each simple conditional density through MDN. It provides a Gaussian Mixture approximation for each simple density, thus introducing assumptions about data. Instead, our approach focuses on continuous multi-dimensional data without taking any a priori data bias.

	A unique work combining DL and non-parametric inference was done by Baird et al.~\cite{Baird05ijcnn}. The authors represent a target pdf via Jacobian determinant of a bijective NN that has an implicit property of non-negativity and total integral to be 1. Additionally, their \emph{pdf learning algorithm} has similarity to our \emph{pdf loss} described in Section \ref{sec:PromOptim}. Although the authors did not connect their approach to virtual physical forces that are pushing a neural surface, their algorithm can be seen as a simple instance of the more general DeepPDF method that we contribute.

	Additionally, in \cite{Nam15toias} the authors describe a loss for pdf ratio estimation which is highly similar to ours. Although our \emph{pdf loss} was developed independently and through different mathematical concepts, Nam et.al. came to the same conclusion by using theory of density ratio estimation \cite{Sugiyama12book}.
	Thus,  insights from \cite{Nam15toias} are highly valuable and relevant also to our approach. The contributions of both our paper and of \cite{Nam15toias} are primarily different in the focused problem. While in \cite{Nam15toias} the authors estimate a ratio between two pdf densities, we use our loss to infer an entire pdf function which is a more challenging task.

	Further, in \cite{Saremi18arxiv} the authors present a method, Deep Energy Estimator Network (DEEN), to infer energy function of the data pdf - an unnormalized model that is equal to (negative) log probability density function, up to a constant. Based on previous research of score matching \cite{Hyvarinen05mlr} and its connection to denoising autoencoders \cite{Hyvarinen08nc,Raphan11nc}, a new approach \cite{Saremi18arxiv} is presented that uses NN as a model for energy approximation. The applied loss can be shown as a force that shapes the surface (the energy approximation) to be parabolic around each data point $x$, eventually providing final surface that in some sense resembles unnormalized KDE estimation. It is done by optimizing the slope of a surface for each input dimension independently, enforcing the slope around each data point $x$ to be zero at $x$, to be positive at nearby smaller points $x^- < x$ and to be negative at nearby larger points $x^+ > x$. Such scheme will have an optimum when surface is parabolic around $x$. Eventually, since the parabolic shape is enforced around each data point, the final estimated energy function will have a structure very close to the real data density, thus approximating it. Note that since the applied loss deals only with the slope of energy function, the final approximation will be very far from its normalized version - that is, the normalization constant can be arbitrary big, probably depending on the initialization of NN's weights. In contrast, the presented herein method, DeepPDF, provides pdf estimation with an overall integral being very close to the true value 1 since our method forces approximation to be point-wise equal to the real pdf function. Hence, although not explicitly normalized, the pdf model produced by DeepPDF is \emph{approximately} normalized.

	Finally, recently Generative Adversarial Networks (GANs) \cite{Goodfellow14nips, Radford15arxiv, Ledig16arxiv} became popular to generate new data samples (e.g.~photo-realistic images). 
	GAN learns a generative model of data samples, thus implicitly learning also the data distribution. Additionally, this technique is applied for distribution transformation and for feature representation learning of high-dimensional measurements, making it extremely important for image processing problems. In contrast, our DeepPDF method learns data distribution explicitly and in theory can also be used for data generation.
	Nonetheless, there is a strong connection between the herein proposed \emph{pdf loss} and recently developed novel GAN losses in \cite{Mohamed16arxiv, Zhao16arxiv, Mao17iccv, Mroueh17nips, Gulrajani17nips, Arjovsky17arxiv}, which we are going to investigate in our future work.

	\section{Probabilistic surface optimization}
	\label{sec:PromOptim}

	\subsection{Derivation}
	
	In this paper we present an approach to learn an arbitrary probability density function (pdf), without making any assumption on the specific distribution family. Concretely, define a multivariate random variable $X_U \in \RR^n$ whose pdf $\probi{U}{X}$ we want to learn and samples from which are available to us. Define a neural network $f$ of arbitrary structure (see more details below) with weights $\theta$ that gets $X \in \RR^n$ as input and returns a scalar. Our goal is for $f(X;\theta)$ to approximate $\probi{U}{X}$. To that end, we develop a  probabilistic optimization method, named \emph{Probabilistic Surface Optimization} (PSO), that uses stochastic \emph{pdf loss} as described below.

	Both $f(X;\theta)$ and $\probi{U}{X}$ functions can be seen as surfaces in $\RR^{n + 1}$:  $n$ dimensions from $X$ support and one additional \emph{probabilistic dimension} for probability values. We look for an optimization technique with loss (w.r.t. $\theta$) that will force surface $f(X;\theta)$ to be similar to surface $\probi{U}{X}$. Moreover, we look for a loss with the property that given enough data samples from $\probs{U}$ the optimization convergence point $f(X;\theta^*)$ will coincide with $\probi{U}{X}$ for any $X\in\RR^n$. We will derive such an optimization method in several simple steps.

	Our method is based on the main observation that performing optimization for loss $L(\theta, X) = f(X;\theta)$ is identical to pushing the virtual surface $f(X;\theta)$, parametrized by $\theta$, at point $X$ with some point-wise force. Indeed, consider a gradient descent (GD) update of a single optimization step for data sample $X$, with $\theta_i = \theta_{i-1} - \delta \cdot \nabla L$ where $\delta$ is the step size and $\nabla L = \frac{\partial f(X;\theta)}{\partial \theta}$.
	Then the height change (\emph{differential}) of the surface  $f(X';\theta)$ at another point $X' \in \RR^n$ can be approximated via a first-order Taylor expansion as (see derivation in Appendix A)
	\vspace{-3pt}
	%
	\begin{equation}
	df(X')
	\triangleq
	f(X';\theta_i) - f(X';\theta_{i-1})
	=
	- \delta \cdot g(X', X, \theta_{i-1})
	,
	\label{eq:PointChange}
	\end{equation}
	\begin{equation}
	g(X_1, X_2, \theta) \triangleq 
	{\frac{\partial f(X_1;\theta)}{\partial \theta}}^T
	\cdot
	\frac{\partial f(X_2;\theta)}{\partial \theta}
	,
	\label{eq:GradDeinition}
	\end{equation}
	where the \emph{gradient similarity} function $g(X_1, X_2, \theta)$ expresses gradient correlation at two points. Thus, at the point $X$ where GD optimization was performed, the surface change will be $- \delta \cdot g(X, X, \theta_{i-1})$, i.e.~proportional to the L2 norm of $\theta$ gradient at this point. In other words, performing such a GD step can be thought of as pushing the surface \down at point $X$ by $df(X)$. Changing sign $L(\theta, X) = - f(X;\theta)$ will simply change the direction of the \emph{differential}/force, pushing the surface \up. Further, the above first-order Taylor expansion was observed empirically to be very accurate for learning rate $\delta$ below 0.01. During the typical training process the learning rate is less than 0.01 for most of the optimization epochs, thus the Eq.~(\ref{eq:PointChange}) provides a good estimation for the real \emph{differential}.

	Intuitively, when pushing a \emph{physical} surface at point $X$, surface height at another point $X' \neq X$ also changes, according to elasticity and physical properties of the surface. Typically, such change diminishes with larger distance between $X$ and $X'$. We argue that the same is true for the NN: according to  Eq.~(\ref{eq:PointChange}), $f(X';\theta)$ at point $X' \neq X$ also changes due to optimization of loss at $X$, with \emph{differential} at $X'$ being proportional to $g(X', X, \cdot)$. Therefore, $g(X', X, \cdot)$ is responsible for correlation in height change at different points. The \emph{gradient similarity} function $g(X', X, \cdot)$ may be seen as a filter, deciding how the push at point $X$ affects surface at point $X'$.
	In our experiments we saw that typically there is an opposite trend between $g(X', X, \cdot)$ and the Euclidean distance $d(X', X)$. Yet, $g(X', X, \cdot)$ does not go entirely to zero for large $d(X', X)$, suggesting that there exist (small) side influence even between far away points.


	Next, consider an example optimization algorithm where at each iteration we sample $X_U \sim \probs{U}$ and $X_D \sim \probs{D}$, with $\probs{D}$ being some other known distribution (e.g.~Gaussian). In case the loss is
	\begin{equation}
	L(\theta, X_U, X_D) = -f(X_U;\theta) + f(X_D;\theta),
	\label{eq:SimpleLoss}
	\end{equation}
	%
	a typical GD (or any other optimization method) update at each iteration will push \up (along \emph{probabilistic dimension}) the surface $f(X;\theta)$ at point $X_U$ and will push \down the surface at point $X_D$. Such loss, also known as the critic loss of Wasserstein GAN \cite{Arjovsky17arxiv}, is partially reasonable for pdf estimation since also the target pdf function $\probi{U}{X}$ on average returns high values for data coming from distribution $\probs{U}$ and low values for data coming from other distributions. 
	Yet, such simple loss does not converge as we will see below.

	Further,
	since the considered loss in Eq.~(\ref{eq:SimpleLoss}) is probabilistic with \emph{pushed} points $X_U$ and $X_D$ being sampled,
	we can calculate what will be (first-order Taylor approximation) the average differential $\expectv [df(X)]$ at a  specific point $X \in \RR^n$ when performing a single GD step (see Appendix B):
	%
	\begin{equation}
	\expectv [df(X)]
	=
	\delta \cdot
	\int
	[\probi{U}{X'} - \probi{D}{X'}]
	\cdot
	g(X', X, \theta)
	dX'
	.
	\label{eq:PointChangeSimpleLoss}
	\end{equation}
	The above equation can be thought of as a  convolution of the signal $[\probi{U}{X'} - \probi{D}{X'}]$ around point $X$ according to the convolution operator $g(X', X, \theta)$.
	At each such point $X'$ there are two opposite forces, \up force $F_{U}(X') \triangleq \probi{U}{X'}$ and \down force $F_{D} \triangleq \probi{D}{X'}$, that push the surface with total force $F_{T}(X') \triangleq F_{U}(X') - F_{D}(X')$. Note  these two forces express the impact for each of the two terms in  Eq.~(\ref{eq:SimpleLoss}), the \up term $-f(X_U;\theta)$ and the \down term $+ f(X_D;\theta)$.

	Since  $F_{T}(X')$ does not change along the optimization (it does not involve $\theta$), it forces the surface to constantly move which prevents the optimization convergence.
	This can be simply shown in case $g(X', X, \theta)$ is a Dirac delta function as follows (for a  general function $g(X', X, \theta)$ see proof in Appendix C).

	The Dirac delta assumption may be seen as taking model $f(X;\theta)$ to its expressiveness upper bound, where for all points $\{ X \}$ disjoint subsets of $\theta$ are allocated to represent surface around each $X$. This will make $f(X;\theta)$ arbitrary flexible since each surface area can be changed independently of others. Yet, this \emph{expressiveness} limit is not achievable since it will require $\theta$ of infinite size. However, the \emph{expressiveness} assumption can be useful for convergence analysis since if the most flexible model can not converge then for sure the typical approximation model (e.g. neural network) will not converge also.

	Given this assumption, from Eq.~(\ref{eq:PointChangeSimpleLoss}) we will have $\expectv [df(X)] = \delta \cdot [F_{U}(X) - F_{D}(X)]
	\cdot g(X, X, \theta)$. It can be interpreted as the opposite forces $F_{U}$ and $F_{D}$ are pushing the surface $f(X;\theta)$ at each point $X \in \RR^n$, without any side influence between different points. Since both forces are point-wise constant along the optimization process, the surface $f(X;\theta)$ will be pushed asymptotically to an infinite height for points $X$ with $F_U(X) > F_D(X) \! \Longleftrightarrow \! \probi{U}{X} > \probi{D}{X}$; at the rest of the points,  $f(X;\theta)$ will be pushed to a negative infinite height. Clearly, such an optimization algorithm will never converge to anything meaningful, unless properly regularized (e.g. see \cite{Arjovsky17arxiv,Gulrajani17nips}).

	Note that the derived above asymptotic values of $f(X;\theta)$'s outputs, $\pm \infty$, typically cannot be really achieved due to limited expressiveness of a given NN architecture. Instead, usually during the optimization of loss in Eq.~(\ref{eq:SimpleLoss}), the $f(X;\theta)$ is increased to very large positive heights at areas of $\probs{U}$ density modes, and is decreased to very large negative heights at areas of $\probs{D}$ density modes. Such increase/decrease of height is unbounded due to unequal forces as was explained above. Eventually, the optimization fails due to numerical instability that involves too large/small numbers.

	One way to enforce the convergence of simple loss in Eq.~(\ref{eq:SimpleLoss}) is by adapting/changing $\probs{D}$ density along the optimization. Indeed, this is the main idea behind the contrastive divergence (CD) method presented in \cite{Hinton02nc} and further improved in \cite{Liu17arxiv}. In CD, a point $X_D$ is approximately sampled from the current pdf estimation $f(X;\theta)$, by performing Monte Carlo with Langevin dynamics \cite{Hinton02nc} or Stein Variational Gradient Descent (SVGD) \cite{Liu17arxiv,Liu16nips}. As explained above, in case at specific $X \in \RR^n$ we have $F_U(X) > F_D(X) \! \Longleftrightarrow \! \probi{U}{X} > \probi{D}{X}$, the surface $f(X;\theta)$ will be pushed up. It in its turn will increase $\probs{D}$ at $X$ since $\probi{D}{X} \approx f(X;\theta)$. Eventually, such optimization will converge when $\probi{U}{X} = \probi{D}{X} = f(X;\theta)$. A similar idea was also applied in GAN setting in \cite{Kim16arxiv}, where sample $X_D$ is generated by the generator NN and where generator output's density is forced to approximate current pdf estimation $f(X;\theta)$ via KL minimization between two.

	Alternatively, consider the following \emph{pdf loss}:
	\begin{equation}
	L_{pdf}(\theta, X_U, X_D) = \\
	-f(X_U;\theta) \cdot \probi{D}{X_U}
	+
	f(X_D;\theta) \cdot
	\Big[
	f(X_D;\theta)
	\Big]^{mf},
	\label{eq:PDFLoss}
	\end{equation}
	where $\probi{D}{X_U}$ is an analytically calculated pdf value at $X_U$ according to \down distribution $\probs{D}$. We use square brackets $\Big[ \cdot \Big]^{mf}$ to specify  \emph{magnitude function} terms that only produce magnitude coefficients but whose gradient w.r.t. $\theta$ is not calculated (\emph{stop gradient} in Tensorflow \cite{Tensorflow_url}). Such terms do not change the direction of the gradient, affecting only its magnitude.
	
	The loss in Eq.~(\ref{eq:PDFLoss}) is still pushing the surface $f(X;\theta)$ up at points $X_U$ (via term $-f(X_U;\theta) \cdot \probi{D}{X_U}$) and down at points $X_D$ (via term $f(X_D;\theta) \cdot
	\Big[
	f(X_D;\theta)
	\Big]^{mf}$). However, the expected differential $\expectv [df(X)]$ has changed due to the new loss terms (see Appendix D):
	\begin{equation}
	\expectv [df(X)]
	=
	\delta \cdot
	\int
	[F_{U}(X') - F_{D}(X'; \theta)]
	\cdot
	g(X', X, \theta)
	dX'
	,
	\label{eq:PointChangePDFLossExp}
	\end{equation}
	with forces $F_U$ and $F_D$ being changed to:
	%
	%
	%
	%
	%
	%
	\begin{equation}
	F_U(X') =
	\probi{U}{X'} \cdot \probi{D}{X'},
	\label{eq:UpForce}
	\end{equation}
	\begin{equation}
	F_D(X'; \theta) = 
	\probi{D}{X'} \cdot f(X';\theta).
	\label{eq:DownForce}
	\end{equation}
	%
	The above \emph{differential} will become zero only at $F_U(X') = F_D(X'; \theta)$, and this is where GD will converge, using an  appropriate learning rate decay. Note, that unlike loss in Eq.~(\ref{eq:SimpleLoss}), the \down force $F_D$ is now a function of $\theta$ and can adapt along optimization to stabilize the opposite \up force $F_U$.

	The PSO optimization is based on pushing the probabilistic surface, represented by a NN $f(X;\theta)$, through forces $F_U$ and $F_D$ (see also Figure \ref{fig:Overview-a}). It does so by applying the \emph{pdf loss} defined in Eq.~(\ref{eq:PDFLoss}) iteratively, where this loss can also be written in batch mode, with multiple samples from data densities (see Section \ref{sec:Appr}). Additionally, \emph{pdf loss} can be also derived in an alternative way as a sampled approximation of (see Appendix E):
	\vspace{-5pt}
	\begin{equation}
	L(\theta) =
	\expectv_{X \sim \probs{D}}
	\bigg[
	\big[
	\probi{U}{X} - f(X;\theta)
	\big]^2
	\bigg]
	,
	\label{eq:SquarePDFLoss}
	\end{equation}
	similarly to the ratio estimation methods in \cite{Sugiyama12book}. Although such derivation is  faster, it lacks the intuition about the physical forces and the virtual surface. Yet, this intuition is very helpful for the below analysis of learning stability.

	\subsection{PSO Analysis}

	
	Here we analyze PSO convergence by assuming, for simplicity, that $g(X', X, \theta)$ is a Dirac delta function. In other words, we assume that there is no correlation and \emph{side-influence} among pushed points. As explained above, this can also be seen as an \emph{expressiveness} limit assumption for extremely flexible surface $f(X;\theta)$ where a push at $X$ affects only the surface height at $X$. We emphasize this assumption is undertaken only to simplify the converegence analysis, while in our experiments $g(X', X, \theta)$ was not constrained in any way. Nevertheless, the provided results (Section \ref{sec:Exper}) support the below analysis. A more general analysis is part of future research.

	Considering Eqs.~(\ref{eq:PDFLoss})-(\ref{eq:DownForce}),
	we can analyze the  work done by PSO in two different perspectives. First is the typical optimization analysis of loss in Eq.~(\ref{eq:PDFLoss}), while second is the analysis of dynamical physical forces on surface in some physical system using Eqs.~(\ref{eq:UpForce}) and (\ref{eq:DownForce}). Both perspectives are two different sides of the same coin. Below we will use each of these perspectives to analyze the properties of PSO. 

	

	Considering an arbitrary point $X \in \RR^n$ and using a \emph{physical} perspective, the balance between \emph{up} and \emph{down} forces at the point happens when:
	$F_U(X) = F_D(X; \theta) 
	\quad \Longrightarrow \quad
	f(X;\theta) = \probi{U}{X}$.
	Convergence to such balance can be explained as follows. In case $f(X;\theta) < \probi{U}{X}$, we can see from Eqs.~(\ref{eq:UpForce}) and (\ref{eq:DownForce}) that $F_U(X) > F_D(X; \theta)$, so the surface at $X$ will be pushed up by the \emph{pdf loss}. If, on the other hand, $f(X;\theta) > \probi{U}{X}$ will be true, then it applies $F_U(X) < F_D(X; \theta)$ and the opposite will happen, i.e.~the surface will be pushed down. Eventually, when balance $f(X;\theta) = \probi{U}{X}$ is achieved, both \emph{up} and \emph{down} forces will be equal and cancel each other; the surface at this point will not be affected anymore. Thus, PSO will indeed optimize $f(X;\theta)$ to be $\probi{U}{X}$ in a \emph{time limit}.


	Intuitively, considering the above analysis and Eq.~(\ref{eq:PointChangeSimpleLoss}) we would like the \emph{gradient similarity} $g(X', X, \theta)$ to resemble kernel function with bandwidth being close to average distance between training points. In this way at each point $X \in \RR^n$ the surface $f(X;\theta)$ will be modified only by pushes at nearby training points with whom $X$ has similar statistical frequency (pdf value). Yet, in typical NN the \emph{gradient similarity} doesn't have the kernel shape, although it has rough opposite correlation with distance between two points. This in its turn raises an interesting research direction for NN architecture with more kernel-like function $g(X', X, \theta)$. We leave this direction for the future research.

	\subsection{Choosing $\probs{D}$}

	For $\probs{D}$ we can use any candidate distribution that covers the support of a target distribution $\probs{U}$. This can be concluded from the following reasons. Consider points $\{X : \probi{U}{X} > 0 \text{ and }  \probi{D}{X} = 0 \}$. From a \emph{physical} perspective, Eqs.~(\ref{eq:UpForce}) and (\ref{eq:DownForce}) tell us that both $F_U$ and $F_D$ are zeros at such points. From an \emph{optimization} perspective, we can conclude the same by noting that the second term in Eq.~(\ref{eq:PDFLoss}), $f(X_D;\theta) \cdot
	\Big[
	f(X_D;\theta)
	\Big]^{mf}$, will never be activated at such points since $X$ cannot be sampled from \down distribution $\probs{D}$. Moreover, the first term, $-f(X_U;\theta) \cdot \probi{D}{X_U}$, has a zero multiplier inside, thus nullifying it. Therefore, at such points there will be no optimization done and $f(X;\theta)$ typically will be just interpolation of surface heights at points around $X$.
	
	Further, consider points $\{X : \probi{U}{X} = 0 \text{ and }  \probi{D}{X} > 0 \}$. The \up force $F_U$ will not be activated since $X$ cannot be sampled from \up distribution $\probs{U}$. The force $F_D$ will push surface down at such $X$ but using \emph{optimization} perspective we can see that once surface height becomes negative, $f(X;\theta) < 0$, the force $F_D$ will change its direction and push surface up due to sign canceling in \down term $f(X_D;\theta) \cdot
	\Big[
	f(X_D;\theta)
	\Big]^{mf}$ in Eq.~(\ref{eq:PDFLoss}). Thus, for such points $X$ the $f(X;\theta)$ will oscillate around zero and given an appropriate learning rate decay it will eventually converge to zero which is also the value of $\probs{U}$ at this point.
	
	Concluding from above, the only requirement on distribution $\probs{D}$ is to wrap support of $\probs{U}$. Using such $\probs{D}$, the surface $f(X;\theta)$ at points $\{X : \probi{U}{X} > 0 \}$ will converge to $\probi{U}{X}$, and at points $\{X : \probi{U}{X} = 0 \}$ it will converge to zero. Empirically we observed that the above "wrap" requirement on $\probs{D}$ is typically enough for proper learning convergence. However, some other constraints on selecting $\probs{D}$ may also exist. For instance, if there is an area where $\{X : \probi{U}{X} \gg \probi{D}{X} \}$ we may have many samples from $\probs{U}$ but no samples from $\probs{D}$. Then the force balance in this area will not be reached and the surface will be pushed to infinity, causing instability in the entire optimization process. Intuitively, for proper convergence the ratio $\frac{\probi{U}{X}}{\probi{D}{X}}$ should be bounded in some dynamical range. A more detailed investigation is required to clarify this point which we consider as future work.


	Typically limits of $\probs{U}$'s support can be easily derived from the training dataset and a proper candidate for $\probs{D}$ can be picked up. Furthermore, in case we do not know $\probs{U}$'s support range and still want to learn a  target pdf without experiencing optimization divergence, we can use a "support-safe" version of \emph{pdf loss}: 
	\vspace{-5pt}
	\begin{multline}
	L_{pdf}^{sprt\!-\!safe}(\theta, X_U, X_D) = \\
	-f(X_U;\theta) \cdot \probi{D}{X_U}
	\cdot
	\Big[
	sign(P_{max} - f(X_U;\theta))
	\Big]^{mf}
	+
	f(X_D;\theta) \cdot
	\Big[
	f(X_D;\theta)
	\Big]^{mf},
	\label{eq:SupSafePDFLoss}
	\end{multline}
	where the new $sign$ term constrains the maximal height of surface $f(X;\theta)$ to be $P_{max}$. Note that such a  constraint will prevent $f(X;\theta)$ to accurately approximate target $\probs{U}$ at points $\{X : \probi{U}{X} > P_{max} \}$.

	Finally, the density estimation can also be done via a variant of the \emph{pdf loss} without the \emph{magnitude function} terms:
	\begin{equation}
	L_{pdf}(\theta, X_U, X_D) = 
	-f(X_U;\theta) \cdot \probi{D}{X_U}
	+ 
	\half
	\Big[
	f(X_D;\theta)
	\Big]^{2},
	\label{eq:PDFLossNoSG}
	\end{equation}
	which has exactly the same optimization gradient as the standard \emph{pdf loss} from Eq.~(\ref{eq:PDFLoss}). 

	\section{Deep PDF approach}
	\label{sec:Appr}

	As explained above, the described PSO and \emph{pdf loss} $L_{pdf}(\theta, X_U, X_D)$ will force a NN $f(X;\theta)$ to approximate $\probi{U}{X}$.
	We can also use PSO in batch regime by sampling sets of points from distributions $\probs{U}$ and $\probs{D}$, $B^U \triangleq \{X_U^i\}_{i = 1}^{N}$ and $B^D \triangleq \{X_D^i\}_{i = 1}^{N}$, and updating $\theta$ appropriately by minimizing loss $L_{T}(\theta) = \frac{1}{N} \sum_{i = 1}^{N} L_{pdf}(\theta, X_U^i, X_D^i)$. Using batch of points for a single $\theta$ update can be seen as many points on the surface $f(X;\theta)$ being pushed \up/\down in order to reach a force balance. Such an update should be much more accurate since the number of points in batch will reduce the stochastic nature of the \emph{pdf loss}, and make real \emph{differential} $df(X)$ closer to its expected value in Eq.~(\ref{eq:PointChangePDFLossExp}). The entire batch learning algorithm, named DeepPDF, is detailed more scrupulously in Appendix F.
	Below, we provide details on different deep learning aspects of the presented approach.

	\paragraph{Learning rate and convergence}

	In our experiments we saw that convergence to the target pdf happens very fast, requiring us only to apply "good" learning rate decay policy. Empirically observed, during the first several thousand iterations the surface $f(X;\theta)$ reaches high similarity with surface $\probi{U}{X}$, and its heights at various points continue to oscillate near the ground truth heights of target surface. Such oscillation around the target surface can be explained by the fact that at each iteration only $N$ points are pushed up/down. These $N$ points are sampled at each iteration from $\probs{U}$/$\probs{D}$ and \up/\down forces are applied only at them during this iteration. Thus, considering only one iteration the true forces are very stochastic and may be far from their averaged expressions in Eqs.~(\ref{eq:UpForce}) and (\ref{eq:DownForce}), but still are not too far since  $f(X;\theta)$ becomes similar to the target. 
	Intuitively, we can see that the proper thing to do is to wait till $f(X;\theta)$ starts to oscillate around some surface and then slowly decrease the learning rate $\delta$. Such a heuristic policy can be achieved for example by exponential decay learning rate which at first stage is constant letting $f(X;\theta)$ to get closer to $\probi{U}{X}$, sequentially reducing learning rate to its minimal value $\delta_{min}$ via 
	$\delta = a \cdot
	b^{floor(t / s)} + \delta_{min}$,
	where $a$ is the start learning rate, $b$ is decay rate, $s$ is number of steps before each learning rate reduction and $t$ is iteration index. We used Adam optimizer \cite{Kingma14arxiv} with start learning rate $a = 10^{-3}$ and decay rate $b = 0.5$. In our experiments we observed that using $s = 200000$ and $\delta_{min} = 10^{-7}$ provides better convergence and higher density estimation accuracy. Allowing PSO to take more time to converge empirically was seen to provide more accurate approximation. Still, it is up to the network architect to define learning rate decay policy according to time vs accuracy preference.
	
	%

	Convergence of \emph{pdf loss} is stochastic as was explained above. Empirically observed, the \emph{pdf loss} from Eq.~(\ref{eq:PDFLoss}) does not provide clear convergence metric since it is too noisy and just oscillates around zero for most part of the optimization. Yet, the value of \emph{pdf loss} without \emph{stop-gradient} (Eq.~(\ref{eq:PDFLossNoSG})) provides better insight about the current optimization convergence, since it has direct connection to the squared error in Eq.~(\ref{eq:SquarePDFLoss}). It reduces monotonically during the optimization and may be used to
	decide when accuracy stops improving (e.g. for \emph{early stopping}).
	Alternatively, it is possible to use learning rate with exponential decay and stop training when learning rate get very close to  $\delta_{min}$
	and $\theta$ does not change anymore.
	Further, value of Eq.~(\ref{eq:PDFLossNoSG}) is also useful for choosing the most accurate model between several trained candidates.

	\paragraph{Overfitting}

	Like any other machine learning method, DeepPDF can overfit to training dataset. This can happen for example, if the used neural network $f(X;\theta)$ is highly expressive and the amount of available points from $\probs{U}$ is small. In such case $f(X;\theta)$ will converge to zero-height surface with peaks at available training points. In a sense, this would not be a mistake since the distribution of such a small and scarce training dataset has indeed flat pdf with peaks around training points. However, in some cases data is very difficult to generate and we need to do the best with a little data that we have. 
	
	In order to detect  overfitting, the typical approach also applicable here is to have a separate validation dataset of data generated from $\probs{U}$, and check its performance (for example, via \emph{pdf loss} in Eq.~(\ref{eq:PDFLossNoSG})). Once validation performance starts to decrease,  overfitting takes place.

	To deal with the overfitting problem, the network architect can decrease expressiveness of the network by reducing its size or adding L2 regularization on weights $\theta$. Empirically we saw that such methods can indeed help. Reducing expressiveness of the network will of course increase the approximation error between $f(X;\theta)$ and $\probi{U}{X}$. Yet, we cannot expect high approximation accuracy when there is only scarce data amount available.

	Still, there is one additional solution to the problem of a little data. We can perform data augmentation to add some diversity noise into our samples, based on our knowledge about the real distribution $\probs{U}$. This technique is very popular in image classification field where it proved to be very useful \cite{Wong16arxiv,Perez17arxiv}, producing object classifier that is robust to various camera light conditions and which is less prone to the overfitting. In case of the density estimation such method will increase variability of sample points and reduce the probability pike around each such point. Yet, the specific data augmentation process will also introduce bias (prior knowledge) in our pdf inference - the properties of the output pdf surface (e.g. its smoothness) will be affected by it. If such prior knowledge is correct and if data is indeed scarce, the data augmentation can tremendously improve accuracy of the density estimation.

	\paragraph{Positive constraint}

	
	Since the network $f(X;\theta)$ is an approximation of a non-negative function $\probi{U}{X}$, it makes sense to explicitly constrain its output to be positive/non-negative. For example, we can use activation functions Relu or Softplus at the end of the network. Alternatively, we can use an exponential function whose outputs are positive. This will be especially appealing since the input of an exponential function will be the log-probability function $\log \probi{U}{X}$. Such log-probability is very useful, for example, in estimation applications.
	
	However, using such activation functions (especially exponential) can damage the numerical stability of  learning. For example, $\exp (a)$ will return infinity for $a > 88$. During training the network signal is not stable and can go above 88 depending on learning rate, causing training process to fail. Such a problem can be dealt with by initializing $\theta$ weights in a different way or by reducing learning rate. Yet, such handling is time-consuming and by itself may slow down the learning process.
	
	Alternatively, we can use $f(X;\theta)$ with an  unconstrained output. As was explained in Section \ref{sec:PromOptim}, the \emph{pdf loss} will eventually force it to be non-negative at points sampled from both $\probs{U}$ and $\probs{D}$. Of course a little oscillation around zero may occur. Also, at other points, where $\{X : \probi{U}{X} = 0 \text{ and }  \probi{D}{X} = 0 \}$, the $f(X;\theta)$ may return negative values since these areas are not affected by PSO optimization (at least not directly). But this can be easily handled by a proxy pdf function:
	\begin{equation}
	\bar{f}(X;\theta) = 
	\begin{cases}
	0,& \text{if } f(X;\theta) < 0 \text{ or } \probi{D}{X} = 0\\
	f(X;\theta),              & \text{otherwise}
	\end{cases}
	\label{eq:PDFProxy}
	\end{equation}

	\paragraph{Inner network architecture}

	Expressiveness of a NN $f(X;\theta)$, i.e.~how well it can approximate a specific surface $\probi{U}{X}$, depends on its inner structure. Number, type and size of inner layers, all these will eventually affect $f(X;\theta)$'s elasticity and the final approximation accuracy. Yet, while the optimal inner architecture of the network depends on the target pdf $\probi{U}{X}$ and its surface properties, in Section \ref{sec:Exper} we show that even a relatively simple network can approximate various target pdfs with high accuracy, being competitive to the state-of-the-art KDE methods. Still, finding an inner architecture that works perfectly for all target surfaces is an open research question.

	%
	%
	%
	%
	%
	%
	%
	
	Additionally, we tested the usage of batch-normalization (BN) and of dropout within our FC layers. We saw empirically that BN prevents DeepPDF from learning the  target pdf. It can be explained by the fact that BN changes the distribution of each batch while DeepPDF basically relies on this distribution to approximate the target pdf. Similarly, dropout was seen to damage accuracy of density estimation. Thus, BN and dropout harm performance of our method and should be avoided for proper DeepPDF convergence.

	\paragraph{Batch size vs.~estimation quality}

	As was explained above the batch size $N$ can be interpreted as amount of points on surface where we push \up/\down during a single iteration. The bigger is $N$ and the more points are pushed at once, the closer are the real applied up/down forces to their expected values in Eqs.~(\ref{eq:UpForce}) and (\ref{eq:DownForce}). Thus, bigger $N$ will produce more stable, fast and accurate approximation results, which was also observed empirically. Up to memory resource constraints, we suggest to use large batches for best performance.


	\section{Experimental evaluation}
	\label{sec:Exper}

	We evaluate our DeepPDF approach on three target densities, whose analytical expressions can be found in Appendix H. The inferred pdfs conceptually represent complicated multi-modal and discontinuous densities, which need to be inferred in real world applications (e.g. estimating measurement model in robotics domain).

	
	%
	%

	In all scenarios, network $f(X;\theta)$ contains 3 FC layers of size 1024 with Relu non-linearity, followed by an output FC layer of size 1 with no activation function. Also, we use uniform distribution for $\probi{D}{X}$. During learning the batch size is 1000 and overall size of training samples $X_U \sim \probi{U}{X}$ for all evaluated methods is $10^{8}$.
	The effect of smaller dataset size on the inference accuracy is part of future investigation.
	The learning rate was decayed through the same decay policy for all scenarios, without use of early stopping.
	The overall number of iterations was around $6 \cdot 10^6$ for each scenario.
	Once trained, we use pdf proxy function from
	Eq.~(\ref{eq:PDFProxy}) as our pdf function approximation.
	DeepPDF was trained with Adam optimizer \cite{Kingma14arxiv} using GeForce GTX 1080 Ti GPU card.

	For comparison, we infer the same densities with KDE method, using two KDE implementations: First is KernelDensity provided by sklearn library \cite{Sklearn_KDE_url} and represents a standard KDE approach. Second is fastKDE \cite{OBrien16csda} which is a  state-of-the-art KDE extension that provides better accuracy and computational efficiency. Optimal parameters for Sklearn-KDE's were found via grid search, while fastKDE performs automatic parameter search.
	
	Additionally, we experiment with a simplified version of PSO in Appendix G
	to demonstrate its power to infer probability of a specific single point.

	\subsection{Learning 2D distributions}
	\label{sec:Exper2D}

	%
	%

	\begin{table}
		\caption{L2 performance comparison of DeepPDF vs KDE}
		\label{tbl:Perf}
		\centering
		\begin{tabular}{lllll}
			\toprule
			Method     & Param.Number & \emph{Columns}     & \emph{Cosine} & \emph{RangeMsr} \\
			\midrule
			DeepPDF & $\approx 2.1 \cdot 10^{6}$ & $1.0363 \cdot 10^{-6}$  & $9.6844 \cdot 10^{-7}$  & $3.2383 \cdot 10^{-8}$   \\
			Sklearn-KDE     & $\approx 10^{8}$ & $5.0364 \cdot 10^{-5}$ & $1.5969 \cdot 10^{-5}$ & $1.5011 \cdot 10^{-6}$      \\
			fastKDE     & $\approx 10^{8}$ & $3.7447 \cdot 10^{-5}$ & $1.8374 \cdot 10^{-6}$  & $5.3903 \cdot 10^{-8}$ \\
			\midrule
			ratio "fastKDE / DeepPDF"    & & $\approx 36$ & $\approx 1.9$  & $\approx 1.66$ \\
			\bottomrule
		\end{tabular}
	\end{table}

\begin{figure}
		\centering
		
		\begin{tabular}{cccc}
			
			\subfloat[\label{fig:Res2-a}]{\includegraphics[width=0.22\textwidth]{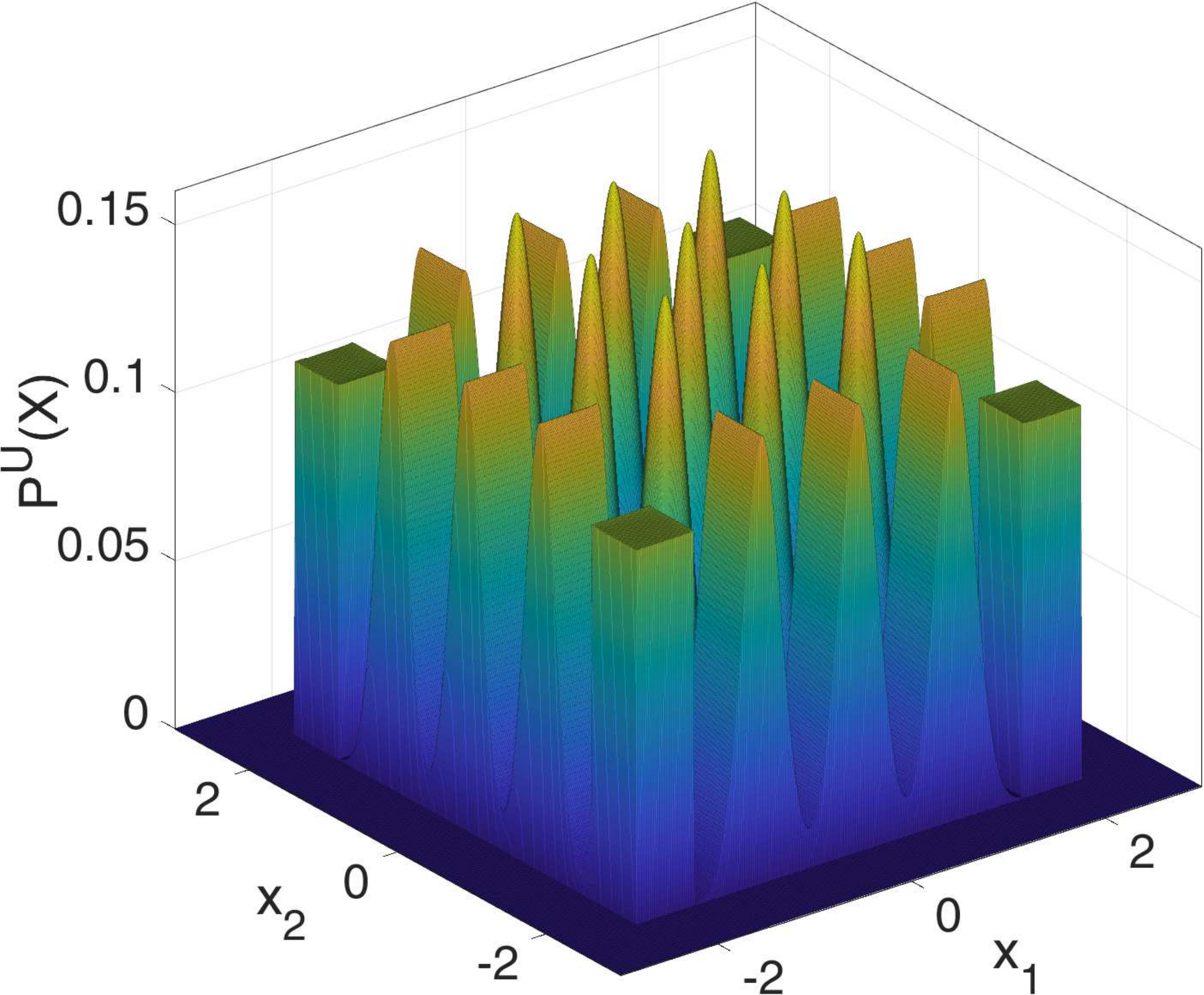}}
			&
			
			\subfloat[\label{fig:Res2-b}]{\includegraphics[width=0.22\textwidth]{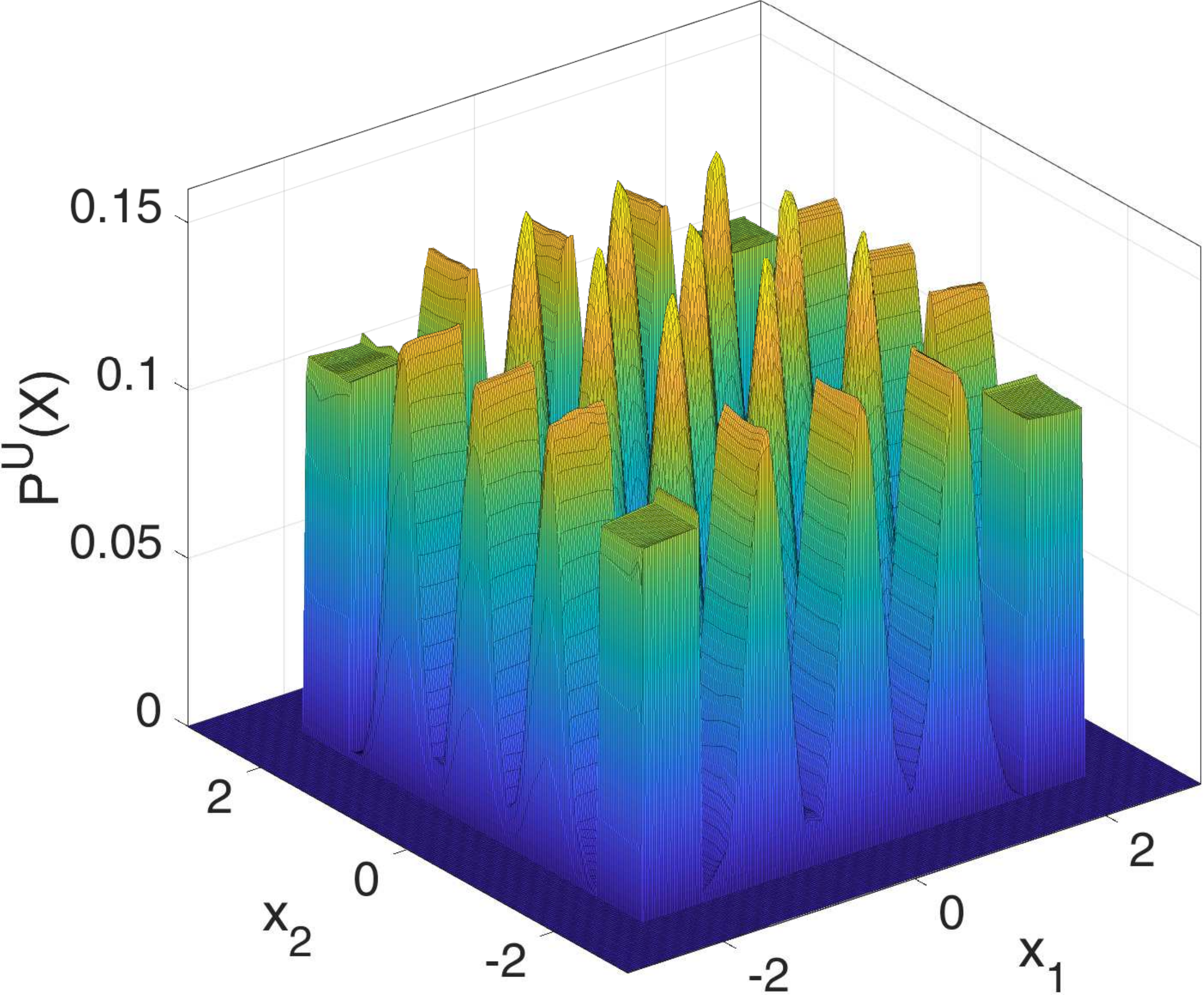}}
			&		
			
			\subfloat[\label{fig:Res2-c}]{\includegraphics[width=0.22\textwidth]{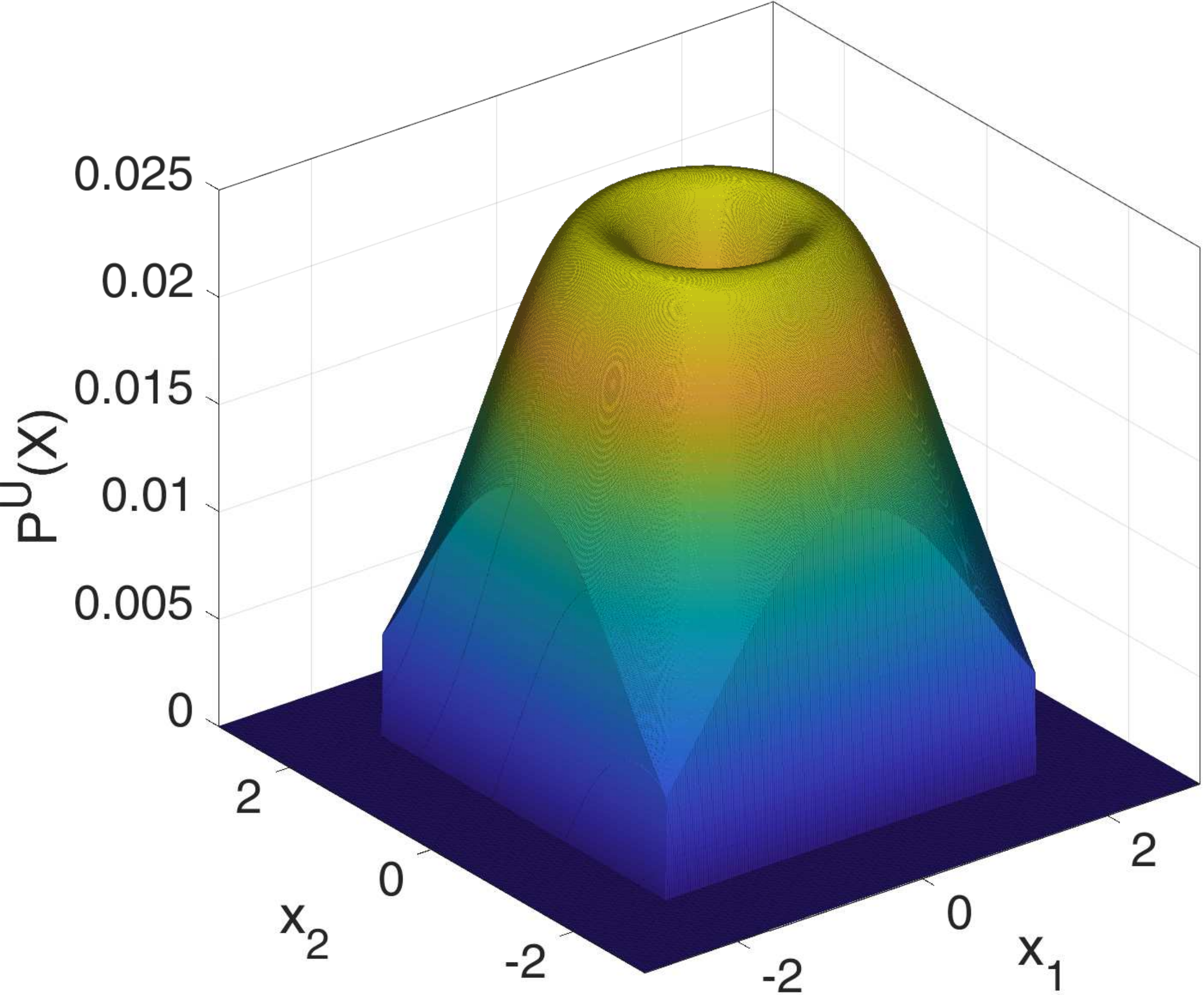}}
			&
			
			\subfloat[\label{fig:Res2-d}]{\includegraphics[width=0.22\textwidth]{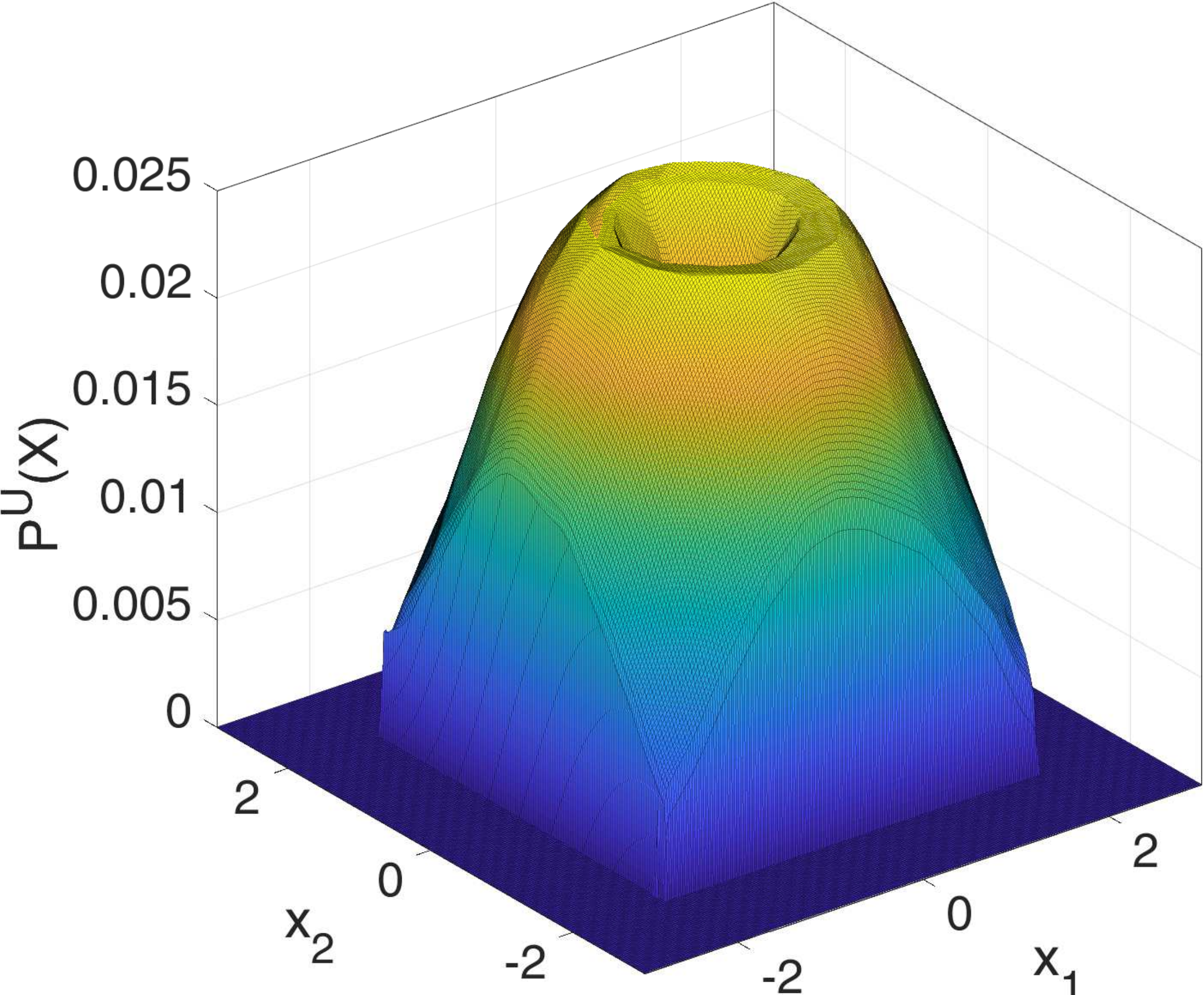}}
			
		\end{tabular}
		
		\protect
		\caption{(a) Example \emph{Columns} target pdf $\probi{U}{X}$ and (b) its DeepPDF estimation. (c) Example \emph{RangeMsr} target pdf $\probi{U}{X}$ and (b) its DeepPDF estimation.
		}
		\label{fig:Res2}
	\end{figure}

	Here we use DeepPDF to estimate density of two 2D distributions shown in Figures \ref{fig:Overview-b} (\emph{Cosine}) and \ref{fig:Res2-a} (\emph{Columns}). The corresponding density approximations are shown in Figures \ref{fig:Overview-c} and \ref{fig:Res2-b}. As can be seen, both distributions were inferred with high accuracy.

	In Table \ref{tbl:Perf} we compare performance of all the three approaches: DeepPDF, Sklearn-KDE and fastKDE. The used performance metric is average L2 distance $\frac{1}{N} \sum_{i = 1}^{N} [ \probi{U}{X_i} - f(X_i) ]^2$, where $f(\cdot)$ is an  approximation function and points $\{ X_i \}_{i=1}^{N}$ form a $257 \times 257$ grid in $\probi{U}{X}$'s support. As can be seen, all methods provide comparable accuracy. For \emph{Columns} pdf the accuracy of DeepPDF is 36 times higher than other techniques, probably because there are density areas with different optimal bandwidths. KDE methods are using the same bandwidth for all inner kernels and cannot approximate such target pdf optimally. However, there is no such problem in DeepPDF since it is highly flexible and capable of approximating a wide variety of functions without the need to specify bandwidth. For \emph{Cosine} pdf the accuracy of fastKDE is only 1.9 times lower than DeepPDF. The reason for such relatively comparable accuracy is the smoothness of the \emph{Cosine} pdf that implies single optimal bandwidth for entire pdf surface.
	

	Also note that query stage of both KDE techniques is very slow - it takes about two days for Sklearn-KDE and about an hour for fastKDE to evaluate 66000 points. In contrast, once trained DeepPDF can evaluate such query number in matter of seconds.
	
	In context of KDE methods, the sample points can be viewed as parameters that represent target pdf approximation function, with such representation being inefficient for the query stage. Thus, the above KDE approaches can be seen as being parametrized by parameter vector of size identical to the size of training dataset ($10^{8}$ in our experiments) multiplied by data dimension. On other hand, the size of representation parameter vector $\theta$ in our DeepPDF method is only about $2.1 \cdot 10^{6}$ ($2103297$ for 2D distributions). Moreover, post-training investigation of $\frac{\partial f(X;\theta)}{\partial \theta}$ on $10^3$ test points showed that only small subset of weights has non-zero gradient on at least one test point. That is, only several tens of thousands of parameters are used to represent surface $f(X;\theta)$, suggesting that most of $\theta$ can be pruned away. Yet, with more compact representation DeepPDF successfully infers the target pdf with even higher accuracy relatively to KDE methods. This supports one of the main arguments of this paper that neural network is more flexible to approximate probability densities than sum of kernels.

	\subsection{Learning 3D range measurement model}
	\label{sec:Exper3D}

	Further, we evaluate the above density techniques to infer a 3D distribution. We use a \emph{RangeMsr} distribution that is similar to the  measurement model of a range sensor which is highly adapted in robotics domain. Each sample point has a structure $[ x, y, f ]$ where $x$ and $y$ are uniformly sampled from range $[-2, 2]$ and symbolically represent robot location. Additionally, $f$ is sampled from $\mathcal{N}(\sqrt{x^2 + y^2}, 1)$ and symbolize distance measurement between robot and some oracle located at origin, plus some noise. Using specific $f = 1$, the \emph{RangeMsr} distribution is visualized in 3D space in Figure \ref{fig:Res2-c}. As can be seen, such distribution has infinitely many modes and approximating it with simple methods like GMM would fail.
	
	In Figure \ref{fig:Res2-d} we can see that DeepPDF's approximation of \emph{RangeMsr} is indeed very accurate. In Table \ref{tbl:Perf} we compare L2 performance of DeepPDF vs KDE methods. As can be seen, our method performs much more accurate estimation than the standard KDE implementation by sklearn library \cite{Sklearn_KDE_url}, and even is superior to a state-of-the-art fastKDE approach \cite{OBrien16csda}.
	

	\section{Conclusions and future work}
	\label{sec:Concl}

	In this paper we presented a novel non-parametric density estimation technique: DeepPDF. It uses a neural network representation of a target pdf which produces high expressiveness, allowing it to approximate an arbitrary multi-dimensional pdf given sample points of thereof. Albeit it requires to design the neural network structure for pdf appropximator, the same simple composition of fully-connected layers with Relu non-linearity was able to approximate all experimented densities with high accuracy. Moreover, once trained, DeepPDF can be efficiently queried for a pdf value at a specific point by a simple network forward pass, making it a much better alternative to the compared KDE methods. We enforce the non-negativity of DeepPDF pdf approximation via proxy function from Eq.~(\ref{eq:PDFProxy}). The total integral is not explicitly constrained to be 1, yet we verified it via numerical integration and saw it to be almost identical to the correct value 1.

	Further, we believe the described probabilistic surface optimization (PSO) is a conceptually novel approach to infer data density and can be applied in other statistical frameworks besides density estimation, such as new data generation. As future research directions, we will investigate both DeepPDF theoretical and empirical convergence properties, and its dependence on data dimension and batch size. Additionally, we will explore the relation between the accuracy of DeepPDF density estimation and size of the training dataset.


\begin{thebibliography}{10}
	
	\bibitem{Antoniak74aos}
	C.~E. Antoniak.
	\newblock Mixtures of {Dirichlet} processes with applications to {Bayesian}
	nonparametric problems.
	\newblock {\em Annals of Statistics}, 2:1152--1174, 1974.
	
	\bibitem{Arjovsky17arxiv}
	Martin Arjovsky, Soumith Chintala, and L{\'e}on Bottou.
	\newblock Wasserstein gan.
	\newblock {\em arXiv preprint arXiv:1701.07875}, 2017.
	
	\bibitem{Baird05ijcnn}
	Leemon Baird, David Smalenberger, and Shawn Ingkiriwang.
	\newblock One-step neural network inversion with pdf learning and emulation.
	\newblock In {\em 2005 IEEE International Joint Conference on Neural Networks,
		IJCNN'05}, volume~2, pages 966--971. IEEE, 2005.
	
	\bibitem{Bengio00nips}
	Yoshua Bengio and Samy Bengio.
	\newblock Modeling high-dimensional discrete data with multi-layer neural
	networks.
	\newblock In {\em Advances in Neural Information Processing Systems (NIPS)},
	pages 400--406, 2000.
	
	\bibitem{Bishop94tr}
	C.M. Bishop.
	\newblock Mixture density networks.
	\newblock Technical report, Aston University, Birmingham, 1994.
	
	\bibitem{Caffe_url}
	{Caffe}.
	\newblock \url{caffe.berkeleyvision.org}.
	
	\bibitem{Duong05sjs}
	Tarn Duong and Martin~L Hazelton.
	\newblock Cross-validation bandwidth matrices for multivariate kernel density
	estimation.
	\newblock {\em Scandinavian Journal of Statistics}, 32(3):485--506, 2005.
	
	\bibitem{Germain15icml}
	Mathieu Germain, Karol Gregor, Iain Murray, and Hugo Larochelle.
	\newblock Made: Masked autoencoder for distribution estimation.
	\newblock In {\em Intl. Conf. on Machine Learning (ICML)}, pages 881--889,
	2015.
	
	\bibitem{Goodfellow14nips}
	Ian Goodfellow, Jean Pouget-Abadie, Mehdi Mirza, Bing Xu, David Warde-Farley,
	Sherjil Ozair, Aaron Courville, and Yoshua Bengio.
	\newblock Generative adversarial nets.
	\newblock In {\em Advances in Neural Information Processing Systems (NIPS)},
	pages 2672--2680, 2014.
	
	\bibitem{Gorur10jcst}
	Dilan G{\"o}r{\"u}r and Carl~Edward Rasmussen.
	\newblock Dirichlet process gaussian mixture models: Choice of the base
	distribution.
	\newblock {\em Journal of Computer Science and Technology}, 25(4):653--664,
	2010.
	
	\bibitem{Gulrajani17nips}
	Ishaan Gulrajani, Faruk Ahmed, Martin Arjovsky, Vincent Dumoulin, and Aaron~C
	Courville.
	\newblock Improved training of wasserstein gans.
	\newblock In {\em Advances in Neural Information Processing Systems (NIPS)},
	pages 5769--5779, 2017.
	
	\bibitem{Heidenreich13asta}
	Nils-Bastian Heidenreich, Anja Schindler, and Stefan Sperlich.
	\newblock Bandwidth selection for kernel density estimation: a review of fully
	automatic selectors.
	\newblock {\em AStA Advances in Statistical Analysis}, 97(4):403--433, 2013.
	
	\bibitem{Hinton06}
	G.E. Hinton, S.~Osindero, and Y.~Teh.
	\newblock A fast learning algorithm for deep belief nets.
	\newblock {\em Neural Computation}, 18:1527--1554, 2006.
	
	\bibitem{Hinton02nc}
	Geoffrey~E Hinton.
	\newblock Training products of experts by minimizing contrastive divergence.
	\newblock {\em Neural Computation}, 14(8):1771--1800, 2002.
	
	\bibitem{Hyvarinen05mlr}
	Aapo Hyv{\"a}rinen.
	\newblock Estimation of non-normalized statistical models by score matching.
	\newblock {\em Journal of Machine Learning Research}, 6(Apr):695--709, 2005.
	
	\bibitem{Hyvarinen08nc}
	Aapo Hyv{\"a}rinen.
	\newblock Optimal approximation of signal priors.
	\newblock {\em Neural Computation}, 20(12):3087--3110, 2008.
	
	\bibitem{Kim16arxiv}
	Taesup Kim and Yoshua Bengio.
	\newblock Deep directed generative models with energy-based probability
	estimation.
	\newblock {\em arXiv preprint arXiv:1606.03439}, 2016.
	
	\bibitem{Kingma14arxiv}
	Diederik~P Kingma and Jimmy Ba.
	\newblock Adam: A method for stochastic optimization.
	\newblock {\em arXiv preprint arXiv:1412.6980}, 2014.
	
	\bibitem{Kopitkov18iros}
	D.~Kopitkov and V.~Indelman.
	\newblock Robot localization through information recovered from cnn
	classificators.
	\newblock In {\em IEEE/RSJ Intl. Conf. on Intelligent Robots and Systems
		(IROS)}. IEEE, October 2018.
	\newblock Accepted.
	
	\bibitem{Krizhevsky12nips}
	Alex Krizhevsky, Ilya Sutskever, and Geoffrey~E Hinton.
	\newblock Imagenet classification with deep convolutional neural networks.
	\newblock In {\em Advances in neural information processing systems}, pages
	1097--1105, 2012.
	
	\bibitem{Larochelle11aistats}
	Hugo Larochelle and Iain Murray.
	\newblock The neural autoregressive distribution estimator.
	\newblock In {\em Proceedings of the Fourteenth International Conference on
		Artificial Intelligence and Statistics}, pages 29--37, 2011.
	
	\bibitem{Ledig16arxiv}
	Christian Ledig, Lucas Theis, Ferenc Husz{\'a}r, Jose Caballero, Andrew
	Cunningham, Alejandro Acosta, Andrew Aitken, Alykhan Tejani, Johannes Totz,
	Zehan Wang, et~al.
	\newblock Photo-realistic single image super-resolution using a generative
	adversarial network.
	\newblock {\em arXiv preprint}, 2016.
	
	\bibitem{Liu16nips}
	Qiang Liu and Dilin Wang.
	\newblock Stein variational gradient descent: A general purpose bayesian
	inference algorithm.
	\newblock In {\em NIPS}, pages 2378--2386, 2016.
	
	\bibitem{Liu17arxiv}
	Qiang Liu and Dilin Wang.
	\newblock Learning deep energy models: Contrastive divergence vs. amortized
	mle.
	\newblock {\em arXiv preprint arXiv:1707.00797}, 2017.
	
	\bibitem{MacEachern98}
	S.~N. MacEachern and P.~Muller.
	\newblock Estimating mixture of dirichlet process models.
	\newblock {\em Journal of Computational and Graphical Statistics}, 7:223--238,
	1998.
	
	\bibitem{Mao17iccv}
	Xudong Mao, Qing Li, Haoran Xie, Raymond~YK Lau, Zhen Wang, and Stephen~Paul
	Smolley.
	\newblock Least squares generative adversarial networks.
	\newblock In {\em Intl. Conf. on Computer Vision (ICCV)}, pages 2813--2821.
	IEEE, 2017.
	
	\bibitem{McLachlan88}
	G.J. McLachlan and K.E. Basford.
	\newblock {\em Mixture Models: Inference and Applications to Clustering}.
	\newblock Marcel Dekker, New York, 1988.
	
	\bibitem{Mohamed16arxiv}
	Shakir Mohamed and Balaji Lakshminarayanan.
	\newblock Learning in implicit generative models.
	\newblock {\em arXiv preprint arXiv:1610.03483}, 2016.
	
	\bibitem{Mroueh17nips}
	Youssef Mroueh and Tom Sercu.
	\newblock Fisher gan.
	\newblock In {\em Advances in Neural Information Processing Systems (NIPS)},
	pages 2510--2520, 2017.
	
	\bibitem{Nam15toias}
	Hyunha Nam and Masashi Sugiyama.
	\newblock Direct density ratio estimation with convolutional neural networks
	with application in outlier detection.
	\newblock {\em IEICE TRANSACTIONS on Information and Systems},
	98(5):1073--1079, 2015.
	
	\bibitem{OBrien16csda}
	Travis~A O’Brien, Karthik Kashinath, Nicholas~R Cavanaugh, William~D Collins,
	and John~P O’Brien.
	\newblock A fast and objective multidimensional kernel density estimation
	method: fastkde.
	\newblock {\em Computational Statistics \& Data Analysis}, 101:148--160, 2016.
	
	\bibitem{Pytorch_url}
	{PyTorch}.
	\newblock \url{pytorch.org}.
	
	\bibitem{Radford15arxiv}
	Alec Radford, Luke Metz, and Soumith Chintala.
	\newblock Unsupervised representation learning with deep convolutional
	generative adversarial networks.
	\newblock {\em arXiv preprint arXiv:1511.06434}, 2015.
	
	\bibitem{Raphan11nc}
	Martin Raphan and Eero~P Simoncelli.
	\newblock Least squares estimation without priors or supervision.
	\newblock {\em Neural Computation}, 23(2):374--420, 2011.
	
	\bibitem{Saremi18arxiv}
	Saeed Saremi, Arash Mehrjou, Bernhard Sch{\"o}lkopf, and Aapo Hyv{\"a}rinen.
	\newblock Deep energy estimator networks.
	\newblock {\em arXiv preprint arXiv:1805.08306}, 2018.
	
	\bibitem{Scott15book}
	David~W Scott.
	\newblock {\em Multivariate density estimation: theory, practice, and
		visualization}.
	\newblock John Wiley \& Sons, 2015.
	
	\bibitem{Sethuraman82sdtrt}
	Jayaram Sethuraman and Ram~C Tiwari.
	\newblock Convergence of dirichlet measures and the interpretation of their
	parameter.
	\newblock In {\em Statistical decision theory and related topics III}, pages
	305--315. Elsevier, 1982.
	
	\bibitem{Silverman18}
	Bernard~W Silverman.
	\newblock {\em Density estimation for statistics and data analysis}.
	\newblock Routledge, 2018.
	
	\bibitem{Sklearn_KDE_url}
	{Sklearn KernelDensity}.
	\newblock
	\url{scikit-learn.org/stable/modules/generated/sklearn.neighbors.KernelDensity.html}.
	
	\bibitem{Smolensky86rep}
	Paul Smolensky.
	\newblock Information processing in dynamical systems: Foundations of harmony
	theory.
	\newblock In D.~E. Rumelhart and J.~L. McClelland, editors, {\em Parallel
		Distributed Processing}, volume~1. The MIT press, Cambridge, MA, 1986.
	
	\bibitem{Sugiyama12book}
	Masashi Sugiyama, Taiji Suzuki, and Takafumi Kanamori.
	\newblock {\em Density ratio estimation in machine learning}.
	\newblock Cambridge University Press, 2012.
	
	\bibitem{Tensorflow_url}
	{TensorFlow}.
	\newblock \url{www.tensorflow.org}.
	
	\bibitem{Uria2013nips}
	Benigno Uria, Iain Murray, and Hugo Larochelle.
	\newblock Rnade: The real-valued neural autoregressive density-estimator.
	\newblock In {\em Advances in Neural Information Processing Systems (NIPS)},
	pages 2175--2183, 2013.
	
	\bibitem{Williams96nc}
	Peter~M Williams.
	\newblock Using neural networks to model conditional multivariate densities.
	\newblock {\em Neural Computation}, 8(4):843--854, 1996.
	
	\bibitem{Zhao16arxiv}
	Junbo Zhao, Michael Mathieu, and Yann LeCun.
	\newblock Energy-based generative adversarial network.
	\newblock {\em arXiv preprint arXiv:1609.03126}, 2016.
	
\end{thebibliography}

\section*{Appendix A: Surface Change after Gradient Descent Update of $L(\theta, X) = f(X;\theta)$}\label{sec:App1}

Performing a single gradient descent step w.r.t. loss $L(\theta, X) = f(X;\theta)$ at a specific \emph{optimized point} $X$ results in a weights update $\theta_i = \theta_{i-1} - \delta \cdot \nabla L$, where $\delta$ is step size and $\nabla L = \frac{\partial f(x;\theta)}{\partial \theta}|_{x = X, \theta = \theta_{i-1}}$. After such update, the surface height at point $X$, $f(X;\theta)$, can be approximated via a Taylor expansion as:
\begin{multline}
f(X;\theta_i) =
f(X;\theta_{i-1})
+
(\theta_i - \theta_{i-1})^T
\cdot
\frac{\partial f(x;\theta)}{\partial \theta}|_{x = X, \theta = \theta_{i-1}}
+\\
+
\half \cdot
(\theta_i - \theta_{i-1})^T
\cdot H_{\theta}
\cdot
(\theta_i - \theta_{i-1})
+
\cdots
=\\
=
f(X;\theta_{i-1})
-
\delta
\cdot
(\frac{\partial f(x;\theta)}{\partial \theta}|_{x = X, \theta = \theta_{i-1}})^T
\cdot
\frac{\partial f(x;\theta)}{\partial \theta}|_{x = X, \theta = \theta_{i-1}}
+\\
+
\half \delta^2 \cdot
(\frac{\partial f(x;\theta)}{\partial \theta}|_{x = X, \theta = \theta_{i-1}})^T
\cdot H_{\theta}
\cdot
(\frac{\partial f(x;\theta)}{\partial \theta}|_{x = X, \theta = \theta_{i-1}})
+
\cdots
,
\label{eq:PointChangeDerivation}
\end{multline}
where:
\begin{equation}
H_{\theta}
\triangleq
\frac{\partial^2 f(x;\theta)}{\partial \theta^2}|_{x = X, \theta = \theta_{i-1}}
.
\label{eq:HessDef}
\end{equation}
Above we can see the first two elements of Taylor expansion where the second element depends on a quadratic step size $\delta^2$. The (learning) step size is typically less than one. Further, assuming it is very small number ($\delta \ll 1$), $f(X;\theta_i)$ can be approximated good enough by a first-order Taylor expansion as:
\begin{equation}
f(X;\theta_i) = f(X;\theta_{i-1})
-
\delta
\cdot
(\frac{\partial f(x;\theta)}{\partial \theta}|_{x = X, \theta = \theta_{i-1}})^T
\cdot
\frac{\partial f(x;\theta)}{\partial \theta}|_{x = X, \theta = \theta_{i-1}}
,
\label{eq:PointChangeDerivation2}
\end{equation}
and define
\begin{equation}
df(X)
\triangleq
f(X;\theta_i) - f(X;\theta_{i-1})
=
- \delta \cdot g(X, X, \theta_{i-1})
,
\label{eq:PointChangeDiff}
\end{equation}
\begin{equation}
g(X_1, X_2, \theta) \triangleq 
{\frac{\partial f(X_1;\theta)}{\partial \theta}}^T
\cdot
\frac{\partial f(X_2;\theta)}{\partial \theta}
,
\label{eq:GradDeinition}
\end{equation}
where $g(\cdot)$ function calculates the \emph{cross-gradient} similarity between two different points $X_1$ and $X_2$, i.e.~the dot product between $\theta$ gradients at these two points. Further, $g(X, X, \theta_{i-1})$ is the L2 norm of $\theta$ gradient at point $X$. 

Similarly, the change of surface height at point $X'$, which is different from the \emph{optimized point} $X$ (i.e.~$X' \neq X$), can be approximated as:
\begin{equation}
df(X')
=
- \delta \cdot g(X', X, \theta_{i-1})
.
\label{eq:Point2ChangeDiff}
\end{equation}
Note that the cross-gradient similarity function is symmetric, i.e.~$g(X', X, \cdot) = g(X, X', \cdot)$. Thus, from the above we can see that for any two arbitrary points $X \in \RR^n$ and $X' \in \RR^n$, pushing at one point will change height at another point according to the cross-gradient similarity $g(X', X, \cdot)$. Therefore, $g(X', X, \cdot)$ is responsible for correlation in height change at different points. Given $|g(X', X, \cdot)|$ is small (or zero), optimizing loss $L$ (pushing surface) at point $X$ will not affect the surface at point $X'$. The opposite is also true: if $|g(X', X, \cdot)|$ is large, then pushing the  surface at point $X$ will have a big impact at point $X'$.

\hfill $\blacksquare$

\section*{Appendix B: Surface Change after Gradient Descent Update of Simple Loss}\label{sec:App2}

In case of simple loss $L(\theta, X_U, X_D) = -f(X_U;\theta) + f(X_D;\theta)$ where 
$X_U$ and $X_D$ are sampled from $\probs{U}$ and $\probs{D}$ respectively, consider weights update $\theta_i = \theta_{i-1} - \delta \cdot \nabla L$. Similarly to Eq.~(\ref{eq:PointChangeDerivation}), the change of surface height at some point $X \in \RR^n$ can be approximated as a first-order Taylor expansion:
\begin{equation}
df(X)
=
\delta \cdot g(X_U, X, \theta_{i-1})
- \delta \cdot g(X_D, X, \theta_{i-1})
.
\label{eq:PointChangeDiffWGAN}
\end{equation}
Further, considering the stochastic nature of $X_U$ and $X_D$, the average differential $\expectv [df(X)]$ can be calculated as:
\begin{multline}
\expectv [df(X)]
=
\delta \cdot 
\int
\probi{U}{X'}
\cdot
g(X', X, \theta_{i-1})
dX'
- \\
-
\delta \cdot 
\int
\probi{D}{X'}
\cdot
g(X', X, \theta_{i-1})
dX'
=
\delta \cdot 
\int
[\probi{U}{X'} - \probi{D}{X'}]
\cdot
g(X', X, \theta_{i-1})
dX'
.
\label{eq:PointChangeDiffWGANExp}
\end{multline}
Again, we see that $g(X', X, \cdot)$ has the role of a filter that controls correlations between points. In contrast, the term $[\probi{U}{X'} - \probi{D}{X'}]$ depends only on point $X'$.


\hfill $\blacksquare$

\section*{Appendix C: Proof of Divergence of Simple Loss}\label{sec:App3}

In case of simple loss $L(\theta, X_U, X_D) = -f(X_U;\theta) + f(X_D;\theta)$ the average differential $\expectv [df(X)]$,
\begin{equation}
\expectv [df(X)]
=
\delta \cdot 
\int
[\probi{U}{X'} - \probi{D}{X'}]
\cdot
g(X', X, \theta_{i-1})
dX'
,
\label{eq:PointChangeDiffWGANExp2}
\end{equation}
contains \emph{signal} term $[\probi{U}{X'} - \probi{D}{X'}]$ and \emph{filter} term $g(X', X, \theta_{i-1})$. Since the \emph{signal} term does not depend on $\theta$ and is unchanging along the optimization, there is no stable point for the optimization convergence. Thus, the learning process diverges. This can be proved by contradiction as follows.

Assume, in contradiction, that the above simple loss converges at optimization step $t$ to the loss's local minimum. Then, for the following optimization steps,  $\theta$, and so $g(X', X, \theta)$, will remain (almost) constant - GD will stay in the minimum area. Hence, the integral in Eq.~(\ref{eq:PointChangeDiffWGANExp2}) becomes a constant for each given point $X$, since $\theta$ does not change anymore. This constant is non-zero, unless both \up and \down densities are identical $\probi{U}{X'} = \probi{D}{X'}$; or the learned $f(X;\theta)$ is flat ($g(X', X, \theta) = 0$ for any two points). Yet, such specific cases are degenerate and we exclude them.

Concluding from the above, after optimization step $t$ the $\expectv [df(X)]$ became non-zero constant for each $X$. Therefore, on average the surface height continues to change by a non-zero constant at each point $X$, going towards $\pm \infty$. Yet, such change of surface $f(X;\theta)$ is impossible since $\theta$ is (almost) constant after $t$. We got a contradiction, meaning that the above convergence assumption is wrong; that is, the  simple loss  $L(\theta, X_U, X_D) = -f(X_U;\theta) + f(X_D;\theta)$ cannot converge due to non-zero signal $[\probi{U}{X'} - \probi{D}{X'}]$.

\section*{Appendix D: Surface Change after Gradient Descent Update of PDF Loss}\label{sec:App4}

In case of \emph{pdf loss} $L_{pdf}(\theta, X_U, X_D) = \\
-f(X_U;\theta) \cdot \probi{D}{X_U}
+
f(X_D;\theta) \cdot
\Big[
f(X_D;\theta)
\Big]^{mf}$ where 
$X_U$ and $X_D$ are sampled from $\probs{U}$ and $\probs{D}$ respectively, consider weights update $\theta_i = \theta_{i-1} - \delta \cdot \nabla L$. Similarly to Eq.~(\ref{eq:PointChangeDerivation}), the change of surface height at some point $X \in \RR^n$ can be approximated as first-order Taylor expansion:
\begin{equation}
df(X)
=
\delta 
\cdot \probi{D}{X_U}
\cdot g(X_U, X, \theta_{i-1})
- \delta \cdot
f(X_D;\theta)
\cdot g(X_D, X, \theta_{i-1})
.
\label{eq:PointChangeDiffPDFLoss}
\end{equation}
Further, considering stochastic nature of $X_U$ and $X_D$, the average differential $\expectv [df(X)]$ can be calculated as:
\begin{multline}
\expectv [df(X)]
=
\delta \cdot 
\int
\probi{U}{X'}
\cdot
\probi{D}{X'}
\cdot
g(X', X, \theta_{i-1})
dX'
- \\
-
\delta \cdot 
\int
\probi{D}{X'}
\cdot
f(X';\theta)
\cdot
g(X', X, \theta_{i-1})
dX'
=\\
=
\delta \cdot 
\int
\probi{D}{X'}
\cdot
[\probi{U}{X'} - f(X';\theta)]
\cdot
g(X', X, \theta_{i-1})
dX'
.
\label{eq:PointChangeDiffPDFLossExp}
\end{multline}

\hfill $\blacksquare$

\section*{Appendix E: PDF Loss as Empirical Approximation of Continuous Squared Loss}\label{sec:App5}

Consider a squared loss between NN surface $f(X;\theta)$ and a target function $\probi{U}{X}$, weighted by the pdf $\probs{D}$ of \down distribution:
\begin{equation}
L(\theta) =
\expectv_{X \sim \probs{D}}
\bigg[
\big[
\probi{U}{X} - f(X;\theta)
\big]^2
\bigg]
=
\int 
\probi{D}{X}
\cdot
\big[
\probi{U}{X} - f(X;\theta)
\big]^2
dX
.
\label{eq:SquarePDFLoss}
\end{equation}
Similarly to the ratio estimation methods in \cite{Sugiyama12book}, we can show that the above loss can be minimized by minimizing our \emph{pdf loss}, as follows:
\begin{equation}
L(\theta) =
\int 
\probi{D}{X}
\cdot
\big[
\probi{U}{X}
\big]^2
+
\probi{D}{X}
\cdot
\big[
f(X;\theta)
\big]^2
-
2 \cdot
\probi{D}{X}
\cdot
\probi{U}{X}
\cdot
f(X;\theta)
dX
.
\label{eq:SquarePDFLossDeriv}
\end{equation}
Since the first term doesn't depend on $\theta$, we can rewrite the loss as:
\begin{multline}
L(\theta) =
\int 
\probi{D}{X}
\cdot
\big[
f(X;\theta)
\big]^2
-
2 \cdot
\probi{D}{X}
\cdot
\probi{U}{X}
\cdot
f(X;\theta)
dX
=\\
=
\int 
\probi{D}{X}
\cdot
\big[
f(X;\theta)
\big]^2
dX
-
2 \cdot
\int 
\probi{U}{X}
\cdot
\probi{D}{X}
\cdot
f(X;\theta)
dX
=\\
=
\expectv_{X \sim \probs{D}} [f(X;\theta)^2]
-
2 \cdot
\expectv_{X \sim \probs{U}} 
[\probi{D}{X}
\cdot
f(X;\theta)]
.
\label{eq:SquarePDFLossDeriv2}
\end{multline}
The above can be approximated by $N$ samples $\{ X_U^i \}_{i = 1}^{N}$ from distribution $\probs{U}$ and $N$ samples $\{ X_D^i \}_{i = 1}^{N}$ from distribution $\probs{D}$ as:
\begin{equation}
L(\theta) =
\frac{1}{N}
\sum_{i = 1}^{N}
f(X_D^i;\theta)^2
-
2 \cdot
\frac{1}{N}
\sum_{i = 1}^{N}
\probi{D}{X_U^i}
\cdot
f(X_U^i;\theta)
\label{eq:SquarePDFLossDeriv3}
\end{equation}
where for notation simplicity we can rewrite it as one-sample version:
\begin{equation}
L(\theta) =
f(X_D;\theta)^2
-
2 \cdot
\probi{D}{X_U}
\cdot
f(X_U;\theta)
.
\label{eq:SquarePDFLossDeriv4}
\end{equation}
Next, note that the above loss has the same gradient w.r.t. $\theta$ as the following loss:
\begin{equation}
L(\theta) =
2
\cdot
f(X_D;\theta)
\cdot 
\Big[
f(X_D;\theta)
\Big]^{mf}
-
2 \cdot
\probi{D}{X_U}
\cdot
f(X_U;\theta)
,
\label{eq:SquarePDFLossDeriv5}
\end{equation}
where $\Big[ \cdot \Big]^{mf}$ specifies  \emph{magnitude function} terms that only produce magnitude coefficients but whose gradient w.r.t. $\theta$ is not calculated (\emph{stop gradient} in Tensorflow \cite{Tensorflow_url}).
Thus, both losses are equal in context of the gradient-based optimization.

After changing order of terms in Eq.~(\ref{eq:SquarePDFLossDeriv5}), and deviding by $2$ (e.g. moving it into learning rate), we will finally get the \emph{pdf loss}:
\begin{equation}
L_{pdf}(\theta, X_U, X_D) = \\
-f(X_U;\theta) \cdot \probi{D}{X_U}
+
f(X_D;\theta) \cdot
\Big[
f(X_D;\theta)
\Big]^{mf}
.
\label{eq:PDFLossFinal}
\end{equation}

\hfill $\blacksquare$

\section*{Appendix F: DeepPDF algorithm in more detail}\label{sec:DeepPDFAlgo}

\begin{algorithm}[H]
	\caption{{\tt DeepPDF} learns pdf $\probi{U}{X}$ using only samples from distribution $\probs{U}$.
	} 
	\label{alg:DeepPDFAlgo} 
	\SetKwInput{Initialize}{Initialize}
	\SetKwBlock{AlgoBody}{begin:}{end}
	\SetKwInput{inputs}{Inputs}{}
	\SetKwInput{outputs}{Outputs}{}
	\inputs{
		
		$G^U$ : generator of data from \up distribution $\probs{U}$ (e.g. dataset sampled from this distribution)
		
		$G^D$ : generator of data from \down distribution $\probs{D}$ (e.g. Normal or Uniform sampler)
		
		$N$ : batch size
		
		$f(X;\theta)$ : neural network whose input has dimension of data, and whose output is a scalar
	}
	\outputs{
		$f(X;\theta)$ : neural network approximation of target pdf $\probi{U}{X}$}
	\BlankLine
	
	\AlgoBody{
		\While{not converged}{
			$\{X_U^i\}_{i = 1}^{N}$ $\Longleftarrow$ sample $N$ points from $G^U$
			
			$\{X_D^i\}_{i = 1}^{N}$ $\Longleftarrow$ sample $N$ points from $G^D$
			
			$L_{T}(\theta) = \frac{1}{N} \sum_{i = 1}^{N} L_{pdf}(\theta, X_U^i, X_D^i)$
			
			Perform one optimization update on $\theta$ to minimize: $\min_{\theta} L_{T}(\theta)$ 
			(e.g. Gradient Descent, Adam, BFGS, etc.)
		}
	}

	\BlankLine
\end{algorithm}

\section*{Appendix G: Point optimization experiment}\label{sec:ExperPoint}

\begin{figure}[h]
	\centering
	
	\begin{tabular}{c}
		
		\subfloat{\includegraphics[width=0.5\textwidth]{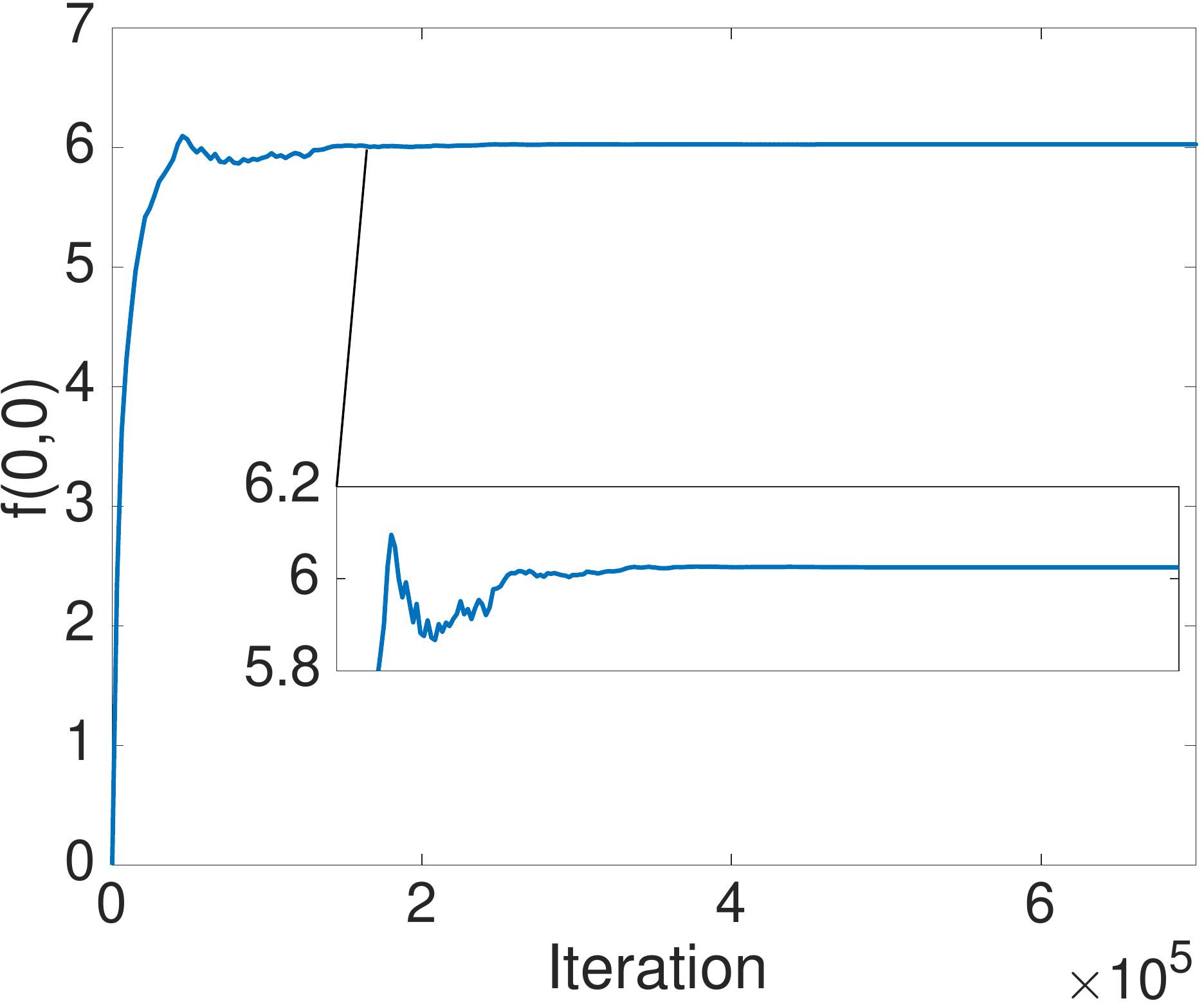}}
		%
		%
		%
	\end{tabular}
	
	\protect
	\caption{Surface height $f(X_z;\theta)$ at $X_z = [0, 0]^T$ during optimization with point loss $L_{zero}(\theta)$ in Eq.~(\ref{eq:PointLoss}).
	}
	\label{fig:Res1}
\end{figure}

In this experiment we apply our loss only at one specific point $X_z = [0, 0]^T$ in $\RR^2$ space, to demonstrate that the described PSO works in a simple case of single point optimization. The used loss is:
\begin{equation}
L_{zero}(\theta) = 
-f(X_z;\theta) \cdot 
\indic_U[X = X_z]
+ 
f(X_z;\theta) \cdot
\Big[
f(X_z;\theta)
\Big]^{mf} \cdot
\indic_D[X = X_z],
\label{eq:PointLoss}
\end{equation}
where $\indic_U[X = X_z]$ is indicator random variable that is equal to 1 with probability $p_U = 0.9$ and is 0 with probability $1 - p_U$. Similarly, $\indic_D[X = X_z]$ is 1 with probability $p_D = 0.15$ and is 0 otherwise. In such a way we simulate a sampling process from  $\probi{U}{X}$ and $\probi{D}{X}$ where we assume $\probi{U}{X_z} = p_U$ and $\probi{D}{X_z} = p_D$. According to the analysis from Section
[3]
, the balance between up/down forces using $L_{zero}(\theta)$ should be at:
\begin{equation}
F_U(X_z) = F_D(X_z) 
\quad \Longrightarrow \quad
p_U = p_D \cdot f(X_z;\theta)
\quad \Longrightarrow \quad
f(X_z;\theta) = 
\frac{p_U}{p_D}
=
6.
\label{eq:BalPointPoint}
\end{equation}
We apply Adam optimizer \cite{Kingma14arxiv} with loss $L_{zero}(\theta)$ during several thousand iterations and show value of $f(X_z;\theta)$ along this optimization in Figure \ref{fig:Res1}. As can be seen, $f(X_z;\theta)$ quickly reaches height 6, following stochastic oscillation around this value. With learning rate decay the oscillations become smaller and the final $f(X_z;\theta)$ value ends to be very close to 6. This simple experiment demonstrates the power of PSO to approximate height of a target pdf. Additionally, it provides insight on how PSO can be used to infer density of discrete random variable instead of continuous, which was the main focus of this paper.

\section*{Appendix H: Used Densities - Equations}\label{sec:App10}

Below are the analytical density functions for distributions used in our experiments.

\begin{enumerate}

	\item{
		\emph{Cosine:}
	}

	\begin{equation}
	\probi{U}{x_1, x_2}
	=
	\frac{1}{\mu}
	\cdot
	\big[
	\cos
	(4
	x_1
	\cdot
	x_2
	)
	+ 1
	\big]
	, \quad
	\mu = 17.631302268269998
	\label{eq:CosineDef}
	\end{equation}

	\item{
		\emph{Columns:}
	}

	\begin{equation}
	\probi{U}{x_1, x_2}
	=
	p(x_1)
	\cdot
	p(x_2)
	,
	\label{eq:ColumnsDef}
	\end{equation}
	where $p(\cdot)$ is mixture distribution with 5 components:\\
	$Uniform(-2.3, -1.7), \mathcal{N}(-1.0, std = 0.2), \mathcal{N}(0.0, std = 0.2), \mathcal{N}(1.0, std = 0.2), Uniform(1.7, 2.3)$\\
	Each component has weight 0.2.

	\item{
		\emph{RangeMsr:}
	}

	\begin{equation}
	\probi{U}{x,y,f}
	=
	p(x) \cdot p(y) \cdot p(f|x,y)
	,
	\label{eq:RangeMsrDef}
	\end{equation}
	where
	\begin{equation}
	p(x)
	=
	p(y)
	=
	Uniform(-2.0, 2.0)
	,
	\label{eq:RangeMsrDef2}
	\end{equation}
	and
	\begin{equation}
	p(f|x,y) = \mathcal{N}(\sqrt{x^2 + y^2}, std = 1.0)
	.
	\label{eq:RangeMsrDef3}
	\end{equation}

\end{enumerate}

\end{document}